\documentclass[lettersize,journal]{IEEEtran}
\usepackage{amsmath,amsfonts}
\usepackage{algorithmic}
\usepackage{algorithm}
\usepackage{array}
\usepackage[caption=false,font=normalsize,labelfont=sf,textfont=sf]{subfig}
\usepackage{booktabs}
\usepackage[switch]{lineno}
\usepackage{multirow}
\usepackage{textcomp}
\usepackage{stfloats}
\usepackage{url}
\usepackage{verbatim}
\usepackage{graphicx}
\usepackage{cite}
\usepackage[table,xcdraw]{xcolor}
\begin{document}
\title{One-Class Risk Estimation for One-Class Hyperspectral Image Classification}

\author{Hengwei~Zhao, Yanfei~Zhong,~\IEEEmembership{Senior~Member,~IEEE,} Xinyu~Wang,~\IEEEmembership{Member,~IEEE,} Hong Shu%

\thanks{This work was supported by the National Natural Science Foundation of China under Grant No.42101327 and No.42071350, in part by the National Key Research and Development Program of China under Grant No. 2022YFB3903502, and LIESMARS Special Research Funding.
\textit{(Corresponding author: Yanfei Zhong)}}
\thanks{Hengwei Zhao, Yanfei Zhong, and Hong Shu are with the State Key Laboratory of Information Engineering in Surveying, Mapping and Remote Sensing, Wuhan University, Wuhan 430079, China (e-mail: whu\_zhaohw@whu.edu.cn; zhongyanfei@whu.edu.cn; shu\_hong@whu.edu.cn).}
\thanks{Xinyu Wang is with the School of Remote Sensing and Information Engineering, Wuhan University, Wuhan 430079, China (e-mail: wangxinyu@whu.edu.cn).}}

\markboth{IEEE Transactions on Geoscience and Remote Sensing}%
{Shell \MakeLowercase{\textit{et al.}}: A Sample Article Using IEEEtran.cls for IEEE Journals}


\maketitle

\begin{abstract}
Hyperspectral imagery~(HSI) one-class classification is aimed at identifying a single target class from the HSI by using only knowing positive data, which can significantly reduce the requirements for annotation.
However, when one-class classification meets HSI, it is difficult for classifiers to find a balance between the overfitting and underfitting of positive data due to the problems of distribution overlap and distribution imbalance.
Although deep learning-based methods are currently the mainstream to overcome distribution overlap in HSI multi-classificaiton, few researches focus on deep learning-based HSI one-class classification. 
In this paper, a weakly supervised deep HSI one-class classifier, namely \emph{HOneCls} is proposed, where a risk estimator---the \emph{One-Class Risk Estimator}---is particularly introduced to make the full convolutional neural network (FCN) with the ability of one class classification in the case of distribution imbalance.
Extensive experiments (20 tasks in total) were conducted to demonstrate the superiority of the proposed classifier.
\end{abstract}


\section{Introduction}
\IEEEPARstart{O}{ne-class} classification, as a weakly supervised leanring problem, is aimed at learning a binary classifier from only knowing positive data~\cite{doi:10.1080/2150704X.2016.1265689}.
Compared with supervised multi-class classification, one-class classification does not require annotating negative data, which can greatly reduce the labeling workload and has broad prospects in variaous earth vision applications~\cite{LEI2021102598,LI2022102947,ZHAO2022328}.
However, few releated works have focus on one-class classification in HSI.

The requirements of only a single target class needing to be identified from the HSI is often raised by the real-world application scenarios.
However, most of the current HSI classification methods focus on the multi-class classification task, where all the classes that exist in the HSI need to be annotated in one-class scenarios.
For example, it would be preferable to only annotate the invasive tree species rather than all the species in the task of invasive tree species mapping in tropical mountain areas, because labeling all the species is labor-intensive and even impractical, due to the high species richness~\cite{PIIROINEN2018119}.
Compared with HSI multi-class classification, the major advantages of HSI one-class classification can be summarized as follows:
\begin{itemize}
\item No predefined class system:
It is difficult to establish a complete class system in the real world, and a feasible way is to define ``all classes except the class of interest as negative"~\cite{10.1145/1401890.1401920}.
\item Only positive annotation:
One-class classification can significantly reduce the requirement for annotation for model training because there is no need to annotate the negative classes.
\end{itemize}

According to the difference of the input data, there are two kinds of one-class classifiers: positive (P) classifiers and positive and unlabeled (PU) classifiers~\cite{doi:10.1080/2150704X.2016.1265689}.
P classifiers, which only use positive data during the training stage, are designed to ``describe'' positive data~\cite{10.1162/NECO_a_00534,10.1145/335191.335388,4276895, 4137865,rs9111161}.
Recent studies have demonstrated that better one-class classification results can be obtained by using extra unlabeled data~\cite{doi:10.1080/2150704X.2016.1265689}.
The two-step PU classification strategy is used in the heuristic methods~\cite{FOODY20061}, which first obtain reliable negative samples from the unlabeled data, and then train a binary classifier by these positive and selected negative samples.
However, the reliability of the selected negative samples seriously affects the final results.
Therefore, the positive data and all the unlabeled data are used simultaneously to train the one-step PU classification model, such as biased classifiers~\cite{PIIROINEN2018119}, post-threshold calibration methods~\cite{5559411,9201373,LU2021112584}, and unbiased risk estimation methods~\cite{LEI2021102598,ZHAO2022328}.
However, the performance of these methods is limited by the characteristics of HSI one-class classification tasks.

\begin{figure}[!t]
\centering
\subfloat[\label{data_distribution}]{\includegraphics[width=0.98\columnwidth]{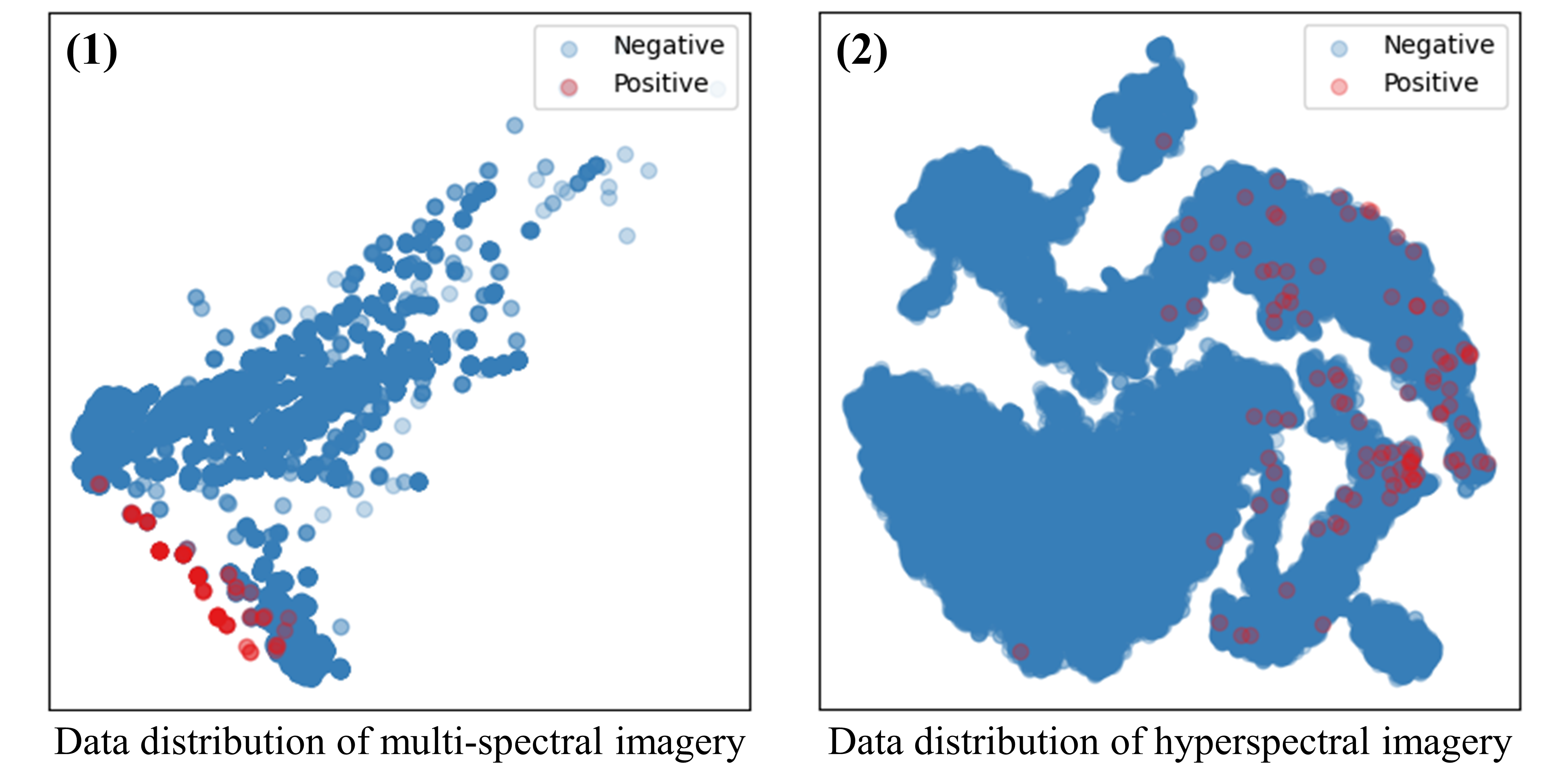}}
\vfill
\subfloat[\label{distribution_imbalance}]{\includegraphics[width=0.98\columnwidth]{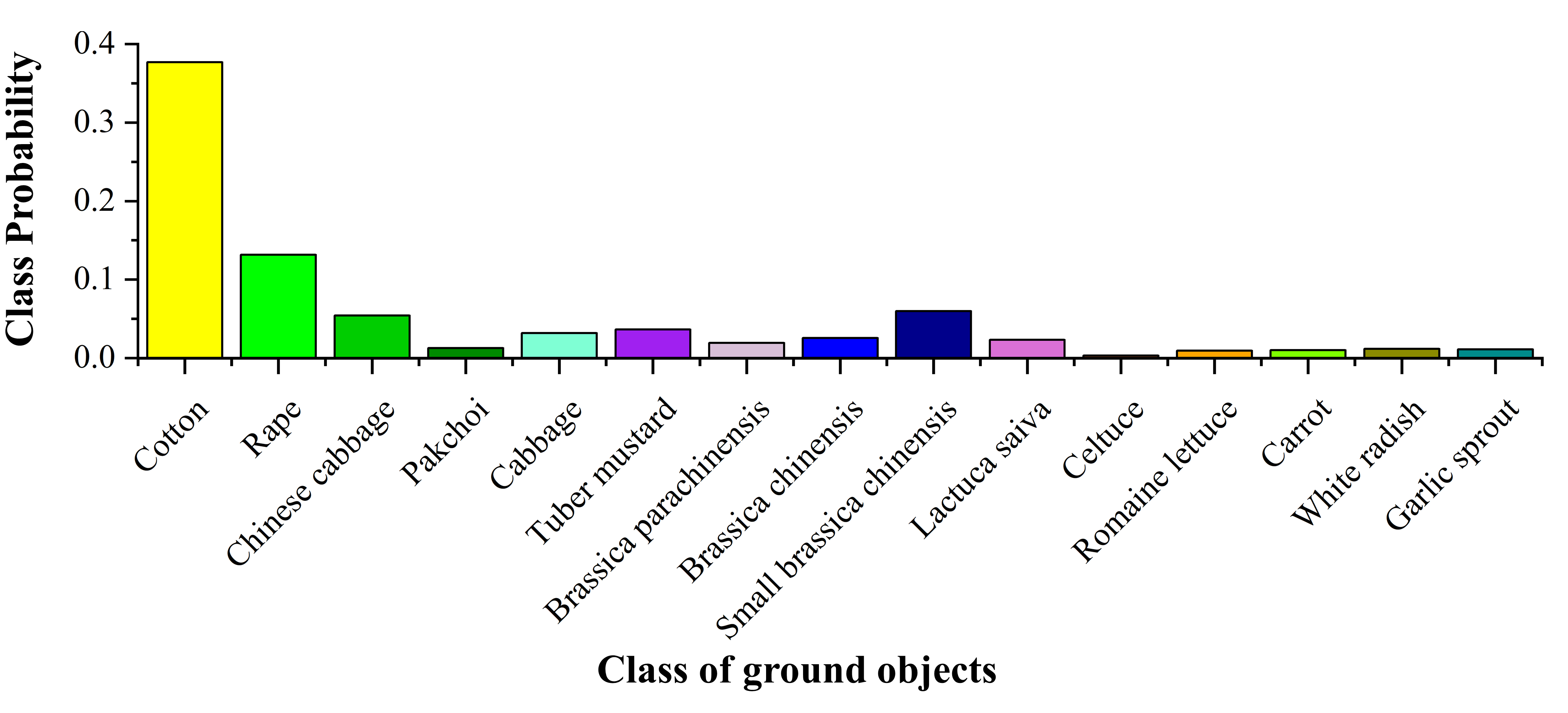}}
\vfill
\subfloat[\label{training_risk}]{\includegraphics[width=0.98\columnwidth]{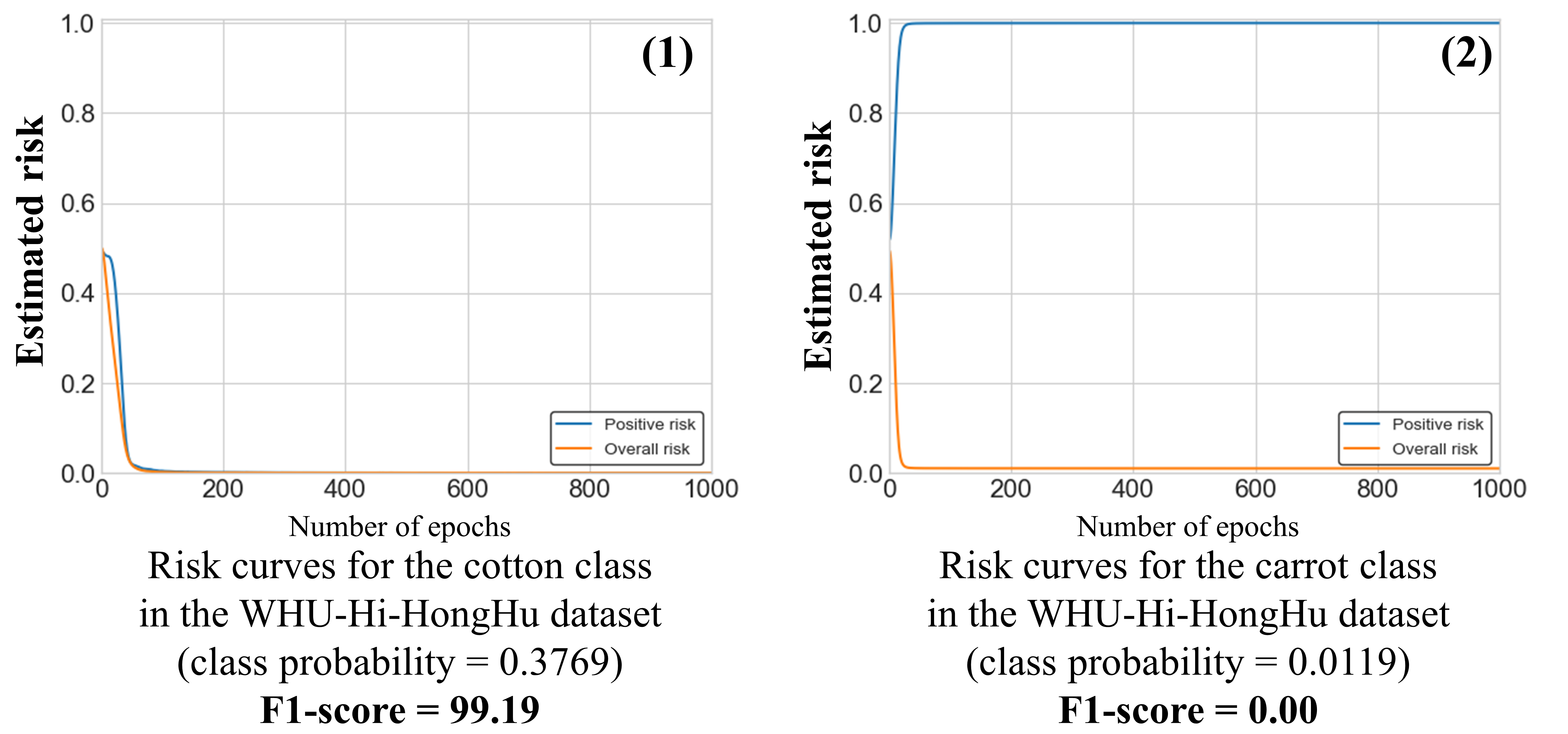}}
\caption{Characteristics of the HSI one-class classification.
(a) Distribution overlap. Take the FanCun multi-spectral dataset~\cite{7486129} (positive class: tree) and HongHu HSI dataset (positive class: tuber mustard) as example. Using tSNE to display the data, the overlap between the distribution of positive and negative HSI data is more severe due to the negative HSI data contain other plants species.
(b) Distribution imbalance. Take the HongHu dataset as an example, the classes of objects and their class probability are shown, the percentage of more than 11 kinds of objects is less than 0.05.
(c) Instance of risk curves of distribution balanced and imbalanced data during training stage (the number of positive training samples for cotton and carrot is fixed at 100)}.
\label{hsi_characteristics}
\end{figure}

As illustrated in~Fig.~\ref{hsi_characteristics}, two properties of HSI one-class classification tasks are demonstrated: distribution overlap and distribution imbalance, which make it difficult for one-class classifiers to find a balance between the overfitting and underfitting of positive data.
In other words, it is difficult to obtain high precision and recall simultaneously for most tasks (more results can be found in Appendix B).
As shown in~Fig.~\ref{data_distribution}, HSI one-class classification tasks are characterized by high overlap of the distribution of the positive and negative data.
Multi-spectral one-class classification methods are often used to mapping coarse categories, such as vegetation.
However, hyperspectral one-class classificaiton methods are often used to mapping fine grained categories, such as a specific plant species and the negative class of HSI usually includes other plant species.
Another characteristic of HSI one-class classification is the abundance of ground objects, which will lead to distribution imbalance.
In distribution imbalanced data, the probability of the positive class, i.e., $P(y=+1)$, is much smaller than the probability of the negative class, i.e., $P(y=-1)$.
The problem of distribution imbalance leads unbiased risk estimation methods underfit the positive training data.
Take the HongHu dataset~\cite{ZHONG2020112012} (Fig.~\ref{distribution_imbalance}) as an example, there are more than 15 types of ground objects, and as the number of ground object categories increases, the percentage of single-class objects will be a smaller value.
During the training stage, the estimated risk of the positive data decreases as the overall data risk decreases in the distribution balanced data, as shown in the Fig~\ref{training_risk}(1), however, the risk of the positive data shows an opposite trend to that of the overall data when distribution imbalance occurs, as shown in the Fig.~\ref{training_risk}(2).
Specifically, distribution imbalance is not class imbalance.
Class imbalance means that the number of positive samples is much smaller than that of negative samples in the training dataset, however, negative data is not utilized in one-class classification.
Distribution imbalance means that the probability of the positive class is much smaller than that of the negative class in the entire HSI, and increasing the number of positive data cannot mitigate the problem of distribution imbalance (more experimental analysis can be found in Section~\ref{sec:results} to demonstrate the difference between class imbalance and distribution imbalance.)

Deep learning based methods is currently the mainstream to overcome the problem of distribution overlap by extracting more robust spectral-spatial features in HSI multi-classification~\cite{9862940,9382264,HU2022147,wang2022cross,liu2023cross,chen2021urban}, however, deep learning-based HSI one-class classification still needs to be studied.
Although post-threshold calibration methods~\cite{5559411,9201373,LU2021112584} and unbiased risk estimation methods~\cite{LEI2021102598,ZHAO2022328} are classifier-independent, satisfactory results cannot be obtained by using deep learning classifiers as the base classifier with these methods.
Post-threshold calibration methods require the base classifier to accurately estimate the posterior probability, which is challenging for deep learning classifiers because they tend to output overconfident predictions~\cite{NEURIPS2019_757f843a}.
The positive risk of deep unbiased risk estimation methods is not decreased in the case of distribution imbalance (Fig.~\ref{training_risk}(2)).
In conclusion, these methods are unable to find a balance between the overfitting and underfitting on positive data and achieve high precision and recall on most tasks simultaneously (more results can be found in Appendix B).

Note that the tasks of HSI one-class classification and HSI target detection are similar: HSI target detection aims to detect targets using target priors~\cite{9400480}, HSI one-class classification aims to classify positive class using known positive samples~\cite{ZHAO2022328}.
However, most HSI target detection methods produce soft-decision maps~\cite{9968036,9400480,9684246,rs11111310,9884362}, while HSI one-class classification methods are expected to produce hard-decision maps to determine categories~\cite{ZHAO2022328,PIIROINEN2018119,LI2022102947}.
Therefore, additional threshold determination methods need to be designed to enable HSI target detection algorithms to be applied to one-class classification tasks and obtain the location of positive objects.

In order to simultaneously overcome the problems of distribution overlap and distribution imbalance in HSI, a deep learning based HSI one-class classifier---\emph{HOneCls}---is proposed.
Specifically, the first negative conformance-based biased risk estimator ---\emph{One-Class Risk Estimator}---is proposed for neural networks.
The main contributions of this paper are summarized as follows:
\begin{itemize}
\item A novel insight into HSI one-class classification is provided, and we showed that the lower weight of the positive risk is the key bottleneck of risk estimation-based methods for learning with distribution overlapped and distribution imbalanced HSI data.
\item A deep weakly supervised classifier---\emph{HOneCls}---is proposed for HSI one-class classification, where the first negative conformance-based biased risk estimator---the \emph{One-Class Risk Estimator}---is particularly proposed to train neural networks without negative data in the case of distribution overlap and distribution imbalance. The consistency of the negative risk estimator in \emph{One-Class Risk Estimator} is proven.
\item Extensive experiments (20 tasks in total) were conducted to demonstrate the superiority of the proposed \emph{HOneCls} in the case of distribution overlapped and distribution imbalanced HSIs.
\end{itemize}

\section{Related Works}
\subsection{Deep Learning Based Hyperspectral Image Classification}
The goal of HSI classification is to assign a semantic label to each HSI pixel~\cite{DING2022246,8824217,9766028,10005113,9775021,9732701}.
Deep neural networks have shown remarkable performances in the HSI classification task.
According to the different learning modes, deep learning based HSI classification methods can be divided into local spatial information learning frameworks and global spectral-spatial feature learning frameworks~\cite{9007624}.

Most of the related studies have been based on local spatial information learning frameworks with the learning target $f_{local}:R^{S{\times}S}{\rightarrow}R$.
The local spatial information learning frameworks first generate $S{\times}S$ HSI patches, \emph{S} is the fixed spatial size, and then the deep neural networks are trained by these patches.
These neural networks include convolutional neural networks (CNNs), generative adversarial networks, stacked autoencoders, recurrent neural networks and so on~\cite{6844831,8356713,8445697,8662780,8661744}, which aim to model the mapping of $f_{local}$.

Recently, some global spectral-spatial feature learning frameworks have been proposed for HSI classification, which consider the HSI classification task as a kind of semantic segmentation task, and aim to model the mapping function $f_{global}:R^{H{\times}W}{\rightarrow}R^{H{\times}W}$ with a fully convolutional neural network (FCN).
A pretrained FCN is utilized in the DMS$^{3}$FE classifier~\cite{7967742} to extract spatial-spectral features.
A two-branch FCN uses a dense conditional random field (CRF) model to balance the local and global information, and a mask matrix is proposed to assist model training in the case of sparse labels for HSI classification~\cite{8737729}.
As a unified patch-free HSI classification framework, the fast patch-free global learning (FPGA) framework~\cite{9007624} utilizes an encoder-decoder based CNN to capture the global spatial information in the HSI.
In addition, a global stochastic stratified sampling strategy is utilized to guarantee the convergence of the network, which means that the FCN can be optimized in an end-to-end manner.

Several methods have been proposed to reduce the number of training samples in HSI classification, including HSI few-shot learning~\cite{9830757} and HSI transfer learning~\cite{9812472,10050427,9540028,9321709}.
However, the objectives of few-shot learning, transfer learning and one-class classification are different.
Few-shot learning aims to train models using a small number of samples from novel classes, and transfer learning aims to leverage knowledge learned from source domain to target domain.
These two methods require a predefined class system, which implies that the model cannot be trained in scenarios where negative data is absent.
Differing from the above HSI classification methods, the classifier proposed in this paper focuses on the weakly supervised one-class classification task.
Furthermore, the proposed classifier can be trained even when negative data is unavailable.

\subsection{One-class Classification}
The target of one-class classification is to learn a binary classifier from only knowing positive data.
The formulation can be described as follows.
The variables of the input space and output space are $X$ and $Y\in\{+1,-1\}$, respectively.
The marginal distribution of the positive and negative classes is recorded as $P_{p}(\boldsymbol{x})=P(\boldsymbol{x}|Y=+1)$ and $P_{n}(\boldsymbol{x})=P(\boldsymbol{x}|Y=-1)$, respectively, and the marginal distribution of the unlabeled data is recorded as $P(\boldsymbol{x})$.
We let $\pi_{p}=P(Y=+1)$ be the class prior probability, which is assumed to be known in most cases, and can be estimated from the PU data~\cite{pmlr-v48-ramaswamy16,NIPS2016_79a49b3e}.
The objective of P classifier is to learn a binary classifier $f$ from $P_{p}(\boldsymbol{x})$, and the objective of PU classifier is to learn a binary classifier $f$ from $P_{p}(\boldsymbol{x})$ and $P(\boldsymbol{x})$.

P classifiers are designed to ``describe'' positive data by information theory~\cite{10.1162/NECO_a_00534}, density estimation~\cite{10.1145/335191.335388}, or geometry~\cite{4276895, 4137865,rs9111161}.
However, it is difficult to find the decision boundary in the case of only positive data, which is the critical limitation of the P classifiers if the target of the task is ``recognition'' or ``classification''~\cite{NEURIPS2018_bd135462}.
In addition, some empirical hyperparameters are needed to balance the degree of overfit and underfit for the positive data~\cite{9201373}.

The two-step strategy is used in the heuristic PU methods, which first obtain reliable negative samples from the unlabeled data, and then train a binary classifier by these positive and selected negative samples, such as S-EM~\cite{10.5555/645531.656022}.
However, the reliability of the selected negative samples seriously affects the final results.
In the one-step PU methods, the positive data and all the unlabeled data are used simultaneously to train the model.
Learning from PU data can be converted to a cost-sensitive problem~\cite{PIIROINEN2018119,1250918}, classifiers post-threshold calibration problem~\cite{5559411,9201373,LU2021112584}, or unbiased risk estimation problem~\cite{LEI2021102598,ZHAO2022328}.
In post-threshold calibration methods, the binary classifiers can be obtained from
\begin{equation}
    f(\boldsymbol{x})=g(\boldsymbol{x})/c,
\label{pul}
\end{equation}
or
\begin{equation}
    f(\boldsymbol{x})=\frac{1-c}{c}\times\frac{g(\boldsymbol{x})}{1-g(\boldsymbol{x})},
\label{pbl}
\end{equation}
where $g(\boldsymbol{x})$ is a classifier that using unlabeled data as negative data to train, and $c=n_p/(n_p+n_u\times\pi_{p})$, $n_p$ is the number of positive data, $n_u$ is the number of unlabeled data.
The unbiased risk estimation-based methods aim to make the risk of the classifiers learned from the PU data equal to that of a Bayesian binary classifier.
The risk of a Bayesian binary classifier can be formulated as:
\begin{equation}
R_{pn}(f)=\mathbb{E}_{(X,Y) \sim P(\boldsymbol{x},y)}[l(f(X),Y)],
\label{bayesian_risk_1}
\end{equation}
where the $l$ is an arbitrary loss function.
However, the performance of these methods is limited by the distribution overlap and distribution imbalance, and using deep learning classifiers as the basic classifier for these methods still faces the problem of overfitting or underfitting positive data.

\section{HOneCls: One-Class Hyperspectral Imagery Classifier}\label{sec:algotithm}
\begin{figure*}[!t]
\centering
\includegraphics[width=0.98\textwidth]{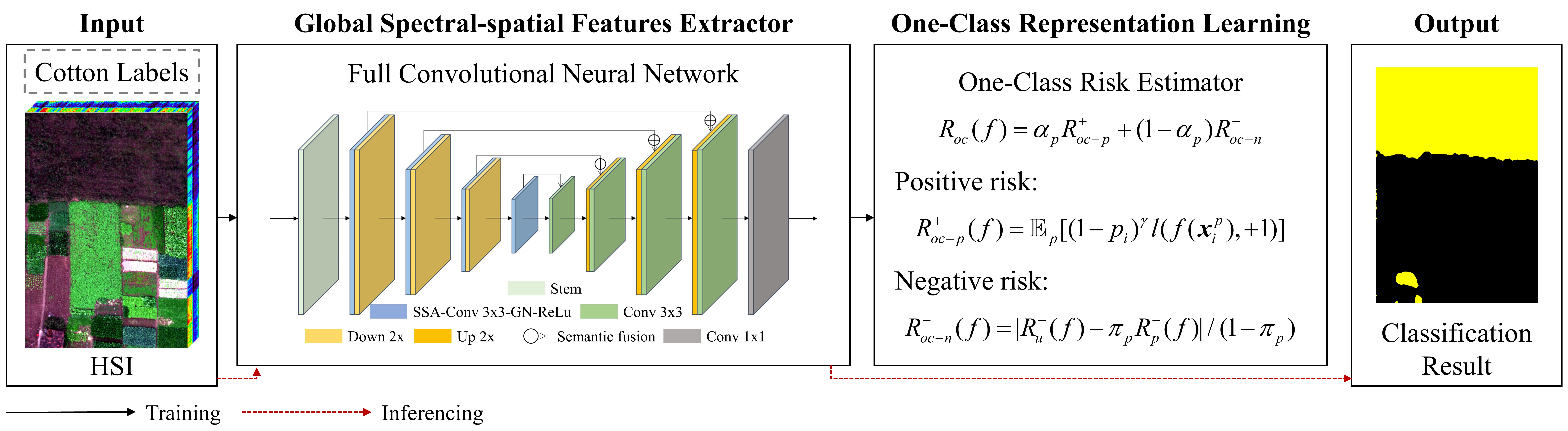}
\caption{Flowchart of the proposed \emph{HOneCls}, which includes the module of global spectral-spatial features extractor and the module of one-class representation learning.
\emph{One-Class Risk Estimator} is responsible for training neural networks to extract robust spectral-spatial features to overcome the problem of distribution overlap in the case of distribution imbalance in one-class classification.}
\label{onecls}
\end{figure*}

In this section, the proposed HSI deep one-class classifier---\emph{HOneCls}---is introduced (Fig.~\ref{onecls}), which includes the module of one-class representation learning and the module of global spectral-spatial features extractor.
\emph{One-Class Risk Estimator} is proposed to train neural networks to extract robust spectral-spatial features to overcome the problem of distribution overlap in the case of distribution imbalance.
Spectral-spatial features can be extracted by any neural network, and a plain FCN---\emph{FreeOCNet} is responsible for extracting global spectral-spatial features in \emph{HOneCls}.

\subsection{One-Class Risk Estimation}
The proposed \emph{One-Class Risk Estimation} includes conformance-based negative risk estimation strategy and positive representation enhancement strategy.
The conformance-based negative risk estimation strategy can be used to estimate a high quality negative risk from neural networks.
The positive representation enhancement strategy can be used to enhance the representation of the positive class to alleviate the problem of distribution imbalance.
Different from unbiased risk estimator, the proposed risk estimator is not equal to Equation~(\ref{bayesian_risk_1}), and extension experiments (20 tasks in total) demonstrate that a one-class classifier also can be obtained from biased risk estimation in the case of conformance based negative risk estimation.

\subsubsection{Conformance Based Negative Risk Estimation Strategy}
The Bayes' theorem is introduced to estimate the risk of negative class without negative data, which means that the expectation of loss of negative data can be estimated without negative data.
As the marginal distribution of the unlabeled samples is equal to the weighted sum of the marginal distributions of the positive and negative classes, i.e., $P(\boldsymbol{x})=\pi_{p}P_{p}(\boldsymbol{x})+(1-\pi_{p})P_{n}(\boldsymbol{x})$, the negative risk can be calculated in the way:
\begin{equation}
\begin{aligned}
 R_{n}^{-}(f) &= \mathbb{E}_{X \sim P(\boldsymbol{x}|y=-1)}[l(f(X),-1)]\\ &= (R_{u}^{-}(f)-{\pi}_{p}R_{p}^{-}(f))/(1-\pi_{p}),   
\end{aligned}
\label{Negative_estimated_risk}
\end{equation}
where $R_{u}^{-}(f)=\mathbb{E}_{u}[l(f(\boldsymbol{x}^u),-1)]$ and $R_{p}^{-}(f)=\mathbb{E}_{p}[l(f(\boldsymbol{x}^p),-1)]$.
The negative risk can be estimated as follows:
\begin{equation}
\widehat{R}_{n}^{-}(f)=(\widehat{R}_{u}^{-}(f)-{\pi}_{p}\widehat{R}_{p}^{-}(f))/(1-\pi_{p}),
\label{Negative_estimated_risk_2}
\end{equation}
where $\widehat{R}^-_u(f)=(1/n_u)\sum_{i=1}^{n_u}l(f(\boldsymbol{x}_i^u),-1)$ and $\widehat{R}^-_p(f)=(1/n_p)\sum_{i=1}^{n_p}l(f(\boldsymbol{x}_i^p),-1)$, and $n_p$ and $n_u$ are the number of PU samples, respectively.
However, the negative risk estimated in this way is low quality and not suitable for neural networks.
Since there is no constraint, $\widehat{R}_{n}^{-}(f)$ will cause the overall risk to become negative, which will result in serious overfitting (more experimental results can be found in Section~\ref{sec:results}).
Formally, the negative risk in the \emph{One-Class Risk Estimator} can be estimated by:
\begin{equation}
\widehat{R}_{oc-n}^-(f)=|\widehat{R}_u^-(f)-\pi_p\widehat{R}_p^-(f)|/(1-\pi_p).
\label{abs_Negative_estimated_risk}
\end{equation}
When $\widehat{R}_{n}^{-}(f)$ is estimated to be negative, $\widehat{R}_{oc-n}^-(f)$ can perform gradient ascent automatically, with the help of the existing deep learning framework, to alleviate the overfitting problem.
${\pi}_{p}\widehat{R}_{p}^{-}(f)$ can also be regarded as an adaptive ``flood level''~\cite{pmlr-v119-ishida20a}, to avoid the overfitting of the positive data in the unlabeled samples.
What is more, the consistency of $\widehat{R}_{oc-n}^-(f)$ are proved as follows, and detailed proof is provided in the Appendix A.

For a fixed $f$, $\widehat{R}_{oc-n}^-(f) \geq \widehat{R}_{n}^-(f)$, this implies the fact that $\widehat{R}_{oc-n}^-(f)$ is biased, but the consistency of $\widehat{R}_{oc-n}^-(f)$ can still be proved.
That is,  $\widehat{R}_{oc-n}^-(f)\to R_{n}^-(f)$ as $n_p, n_u \to \infty$ for a fixed $f$.

\noindent {\bf{Lemma 1.}} The data $(\mathcal{X}_p,\mathcal{X}_u)$ can be divided into two sets, $\mathcal{S}^+(f)=\{(\mathcal{X}_p,\mathcal{X}_u)|\widehat{R}^-_u(f)-\pi_p\widehat{R}^-_p(f) \ge 0\}$ and $\mathcal{S}^-(f)=\{(\mathcal{X}_p,\mathcal{X}_u)|\widehat{R}^-_u(f)-\pi_p\widehat{R}^-_p(f) < 0\}$.
Assuming there are $\mathcal{C}_l$ and $\alpha>0$ such that $0 \le l(f(\boldsymbol{x}),\pm 1) \le \mathcal{C}_l$ and $(1-\pi_p)R_n^-(f) \ge \alpha$, the probability measure of $\mathcal{S}^-(f)$ can be bounded by:
\begin{equation}
\label{lemma1}
P(\mathcal{S}^{-}(f))\le exp(-2(\alpha/\mathcal{C}_l)^2/(\pi_p^2/n_p+1/n_u)).
\end{equation}
The exponential decay of the bias and the consistency of the statistics can be proved based on Lemma 1.
The right-hand side of (\ref{lemma1}) is denoted as $\Delta_f$ in the following.

\noindent {\bf{Theorem 1}} (bias and consistency).
As the number of positive and unlabeled samples increases, $n_p, n_u \rightarrow \infty$, the exponential decay shows up in the bias of $\widehat{R}_{oc-n}^-(f)$:
\begin{equation}
\label{theorem1_bias}
0 \le \mathbb{E}_{\mathcal{X}_p,\mathcal{X}_u}[\widehat{R}_{oc-n}^-(f)]-R_n^-(f) \le \frac{2}{1-\pi_p}\pi_p\mathcal{C}_l\Delta_f.
\end{equation}
Furthermore, we let $\mathcal{X}_{n_p,n_u}=\pi_p/\sqrt{n_p}+1/\sqrt{n_u}, \mathcal{C}_\sigma = \mathcal{C}_l\sqrt{ln(2/\sigma)/2}/(1-\pi_p)$ and for any $\sigma > 0$, with the probability at least $1-\sigma-\Delta_f$,
\begin{equation}
\label{theorem1_consistency}
|\widehat{R}_{oc-n}^-(f)-R_n^-(f)| \le C_\sigma\cdot\mathcal{X}_{n_p,n_u}.
\end{equation}

Equation (\ref{theorem1_consistency}) implies the fact that $\widehat{R}_{oc-n}^-(f)\to R_n^-(f)$ as $n_p, n_u\to \infty$ for a fixed $f$.

\subsubsection{Positive Representation Enhancement Strategy}
In the following, we first analyze the reason for the underfitting of positive data in the case of distribution imbalance: the positive risk is given a lower weight, and then introduce the positive representation enhancement strategy.

From the risk of the Bayesian classifier, the binary classifier $f^{*}(\boldsymbol{x})$ can be obtained by $f^{*}=\mathop{\arg\min}\limits_{f}R_{pn}(f)$.
According to the Bayes' theorem, i.e., $P(\boldsymbol{x})=\pi_{p}P_{p}(\boldsymbol{x})+(1-\pi_{p})P_{n}(\boldsymbol{x})$, Equation~(\ref{bayesian_risk_1}) can be simplified as follows:
\begin{equation}
R_{pn}(f)=\pi_{p}R_{p}^{+}(f)+(1-\pi_{p})R_{n}^{-}(f),
\label{bayesian_risk_2}
\end{equation}
where the risk of the positive data is $R_{p}^{+}(f)=\mathbb{E}_{p}[l(f(\boldsymbol{x}^p),+1)]$ and the risk of the negative data is $R_{n}^{-}(f)=\mathbb{E}_{n}[l(f(\boldsymbol{x}^n),-1)]$.

It is clear from Equation~(\ref{bayesian_risk_2}) that the ratio of the weight of the positive risk and the weight of the negative risk is limited to $\pi_{p}/(1-\pi_{p})=P(Y=+1)/P(Y=-1)$ in unbiased risk estimators, and this ratio does not change with the number of training samples.
However, in order to make the risk estimated from the flexible deep neural network be of high quality, the assumption that there is $\mathcal{C}_l$ such that $\mathop{\max}\limits_{y}l(f(\boldsymbol{x}),y)\leq \mathcal{C}_l$ is made to ensure that the risk estimated from PU data is consistent with $R_{pn}(f)$.
In addition, so that the gradient can be computed everywhere, sigmoid loss $l_{sig}(f(\boldsymbol{x}),y)=1/(1+exp(yf(\boldsymbol{x})))$ is usually chosen as the loss function.
Compared with the cross-entropy loss function, the sigmoid loss is a two-end saturation function, where we let $t=yf(\boldsymbol{x})$, which means that $t \rightarrow -\infty$ will lead to $l^{'}_{sig}(t) \rightarrow 0$ , and $t \rightarrow +\infty$ will lead to $l^{'}_{sig}(t) \rightarrow 0$.

When distribution imbalanced data meets a two-end saturated loss function, the problem of positive class under-fitting may arise.
Based on the observation shown in Fig.~\ref{training_risk}, we found that, although the overall risk is decreasing, the distribution imbalance causes the risk of the positive class to first rise at the beginning of the training.
Due to the utilization of a two-end saturated loss function, when the risk of the positive class reaches its maximum value, the gradient brought by the positive data is a small value, and the gradient from the positive data plays only a small role in the overall gradient, so that the positive data are not fully fitted.

One solution is to enhance the representation of the positive class and accelerate the reduction of positive risk to avoid under-fitting.

The first step in positive representation enhancement is to rebalance the distribution.
Based on Equation~(\ref{bayesian_risk_2}), the ratio of the weight of the positive risk and the weight of the negative risk is limited to $P(Y=+1)/P(Y=-1)$, and this ratio does not change with the number of training samples.
Beyond this ratio, the balancing factor ${\alpha}_{p}{\in}[0,1]$ is introduced in the estimator to rebalance the risk between the positive and negative classes, which aims to balance the distribution of the different classes without changing the positive data distribution and negative data distribution.
Formally, the \emph{One-Class Risk Estimator} is defined as:
\begin{equation}
\widehat{R}_{oc}(f)={\alpha}_{p}\widehat{R}_{oc-p}^{+}(f)+(1-{\alpha}_{p})\widehat{R}_{oc-n}^{-}(f).
\label{Focal_One_Class_Risk_Estimator}
\end{equation}
A naive way is to reweight each positive sample by ${\alpha}_{p}$, however, positive data participate in the process of negative distribution estimation, i.e., $\widehat{R}^-_p(f)$, so this naive way will change the distribution of the negative data in the input space.
The distribution-rebalancing strategy operates on the class level, and does not change the distribution of the positive and negative classes, but enhances the representation of the positive class by adjusting the probability ratio of the positive and negative classes ($P(Y=+1)/P(Y=-1)$) in the input space.

The second step is hard sample mining based positive risk estimation.
Hard sample mining has been widely used in supervised learning; however, due to the lack of negative samples and the positive samples participating in the calculation of negative risk, how to mine hard samples in PU data is not intuitive.
In the setting of learning from PU data, the positive dataset is considered to be a clean dataset, which implies that the positive dataset has no negative samples.
Inspired by Focal Loss~\cite{8417976}, the dynamic modulating factor is only added for each positive sample in the risk estimation of the positive class.
Formally, the risk of the positive class in the \emph{One-Class Risk Estimator} is defined as:
\begin{equation}
R_{oc-p}^{+}(f)=\mathbb{E}_{p}[(1-p_{i})^{\gamma}l(f(\boldsymbol{x}_i^p),+1)],
\end{equation}
which can be estimated as follows:
\begin{equation}
\widehat{R}_{oc-p}^{+}(f)=(1/n_p)\sum_{i=1}^{n_p}(1-p_{i})^{\gamma}l(f(\boldsymbol{x}_i^p),+1),
\label{Focal_positive_risk_estimation}
\end{equation}
where ${\gamma}>0$ is the tunable focusing parameter, and $p_{i}{\in}[0,1]$ implies the confidence of sample $\boldsymbol{x}_i^p$ being judged to be a positive sample, $p_{i}=1/(1+\exp^{-f(\boldsymbol{x}_i)})$.
Two properties are noted in the positive risk estimation.
1)~When sample $\boldsymbol{x}_i^p$ is misclassified, the dynamic modulating factor will be near to 1, and the risk of this sample will be unaffected.
When this sample is well classified, the dynamic modulating factor will be near to 0.
2)~The focusing parameter controls the weight decrease rate of the easy samples.
When ${\gamma}=0$, the positive risk estimation is equivalent to the traditional positive risk estimation.

Finally, the proposed \emph{One-Class Risk Estimator} for HSI one-class classification can be formulated as:
$$\widehat{R}_{oc}(f)={\alpha}_{p}\widehat{R}_{oc-p}^{+}(f)+(1-{\alpha}_{p})\widehat{R}_{oc-n}^{-}(f),$$
where the positive risk $\widehat{R}_{oc-p}^{+}(f)$ can be calculated as shown in (\ref{Focal_positive_risk_estimation}), and the negative risk $\widehat{R}_{oc-n}^{-}(f)$ can be calculated as shown in (\ref{abs_Negative_estimated_risk}).

\begin{figure*}[!t]
    \centering
    \subfloat[\label{HongHu_dataset}]{\includegraphics[width=0.33\textwidth]{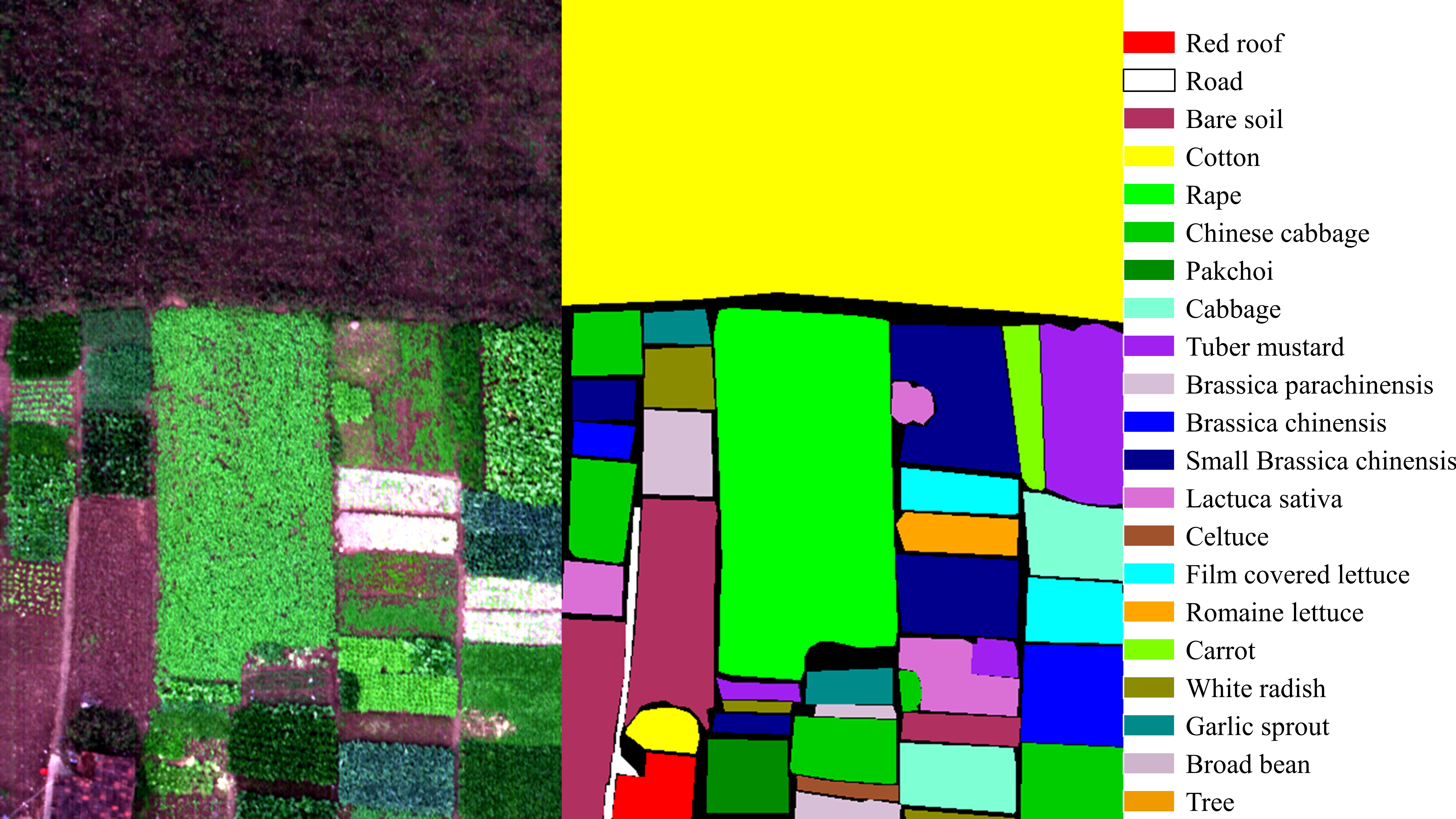}}
    \hfill
    \subfloat[\label{LongKou_dataset}]{\includegraphics[width=0.33\textwidth]{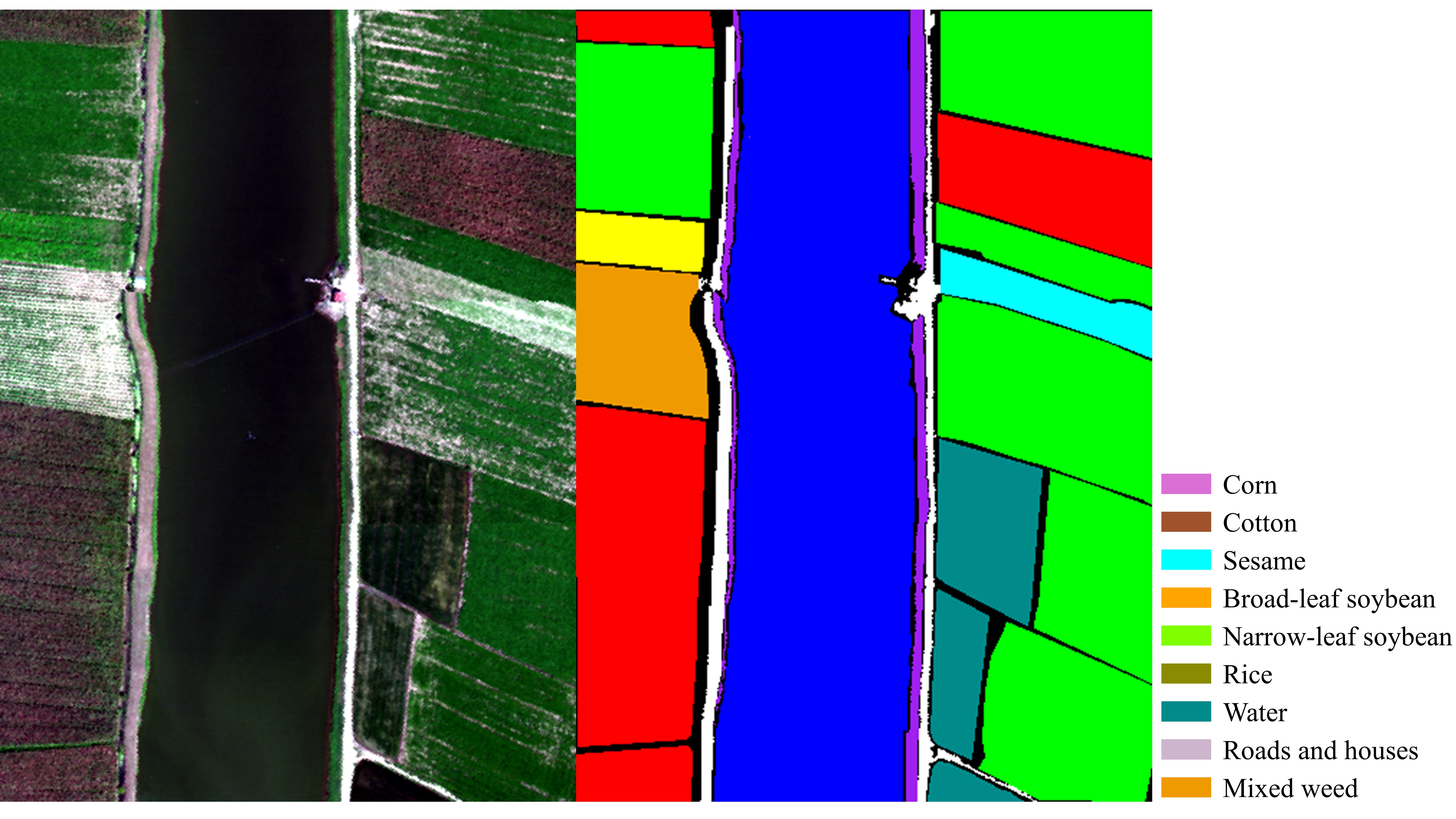}}
    \hfill
    \subfloat[\label{HanChuan_dataset}]{\includegraphics[width=0.33\textwidth]{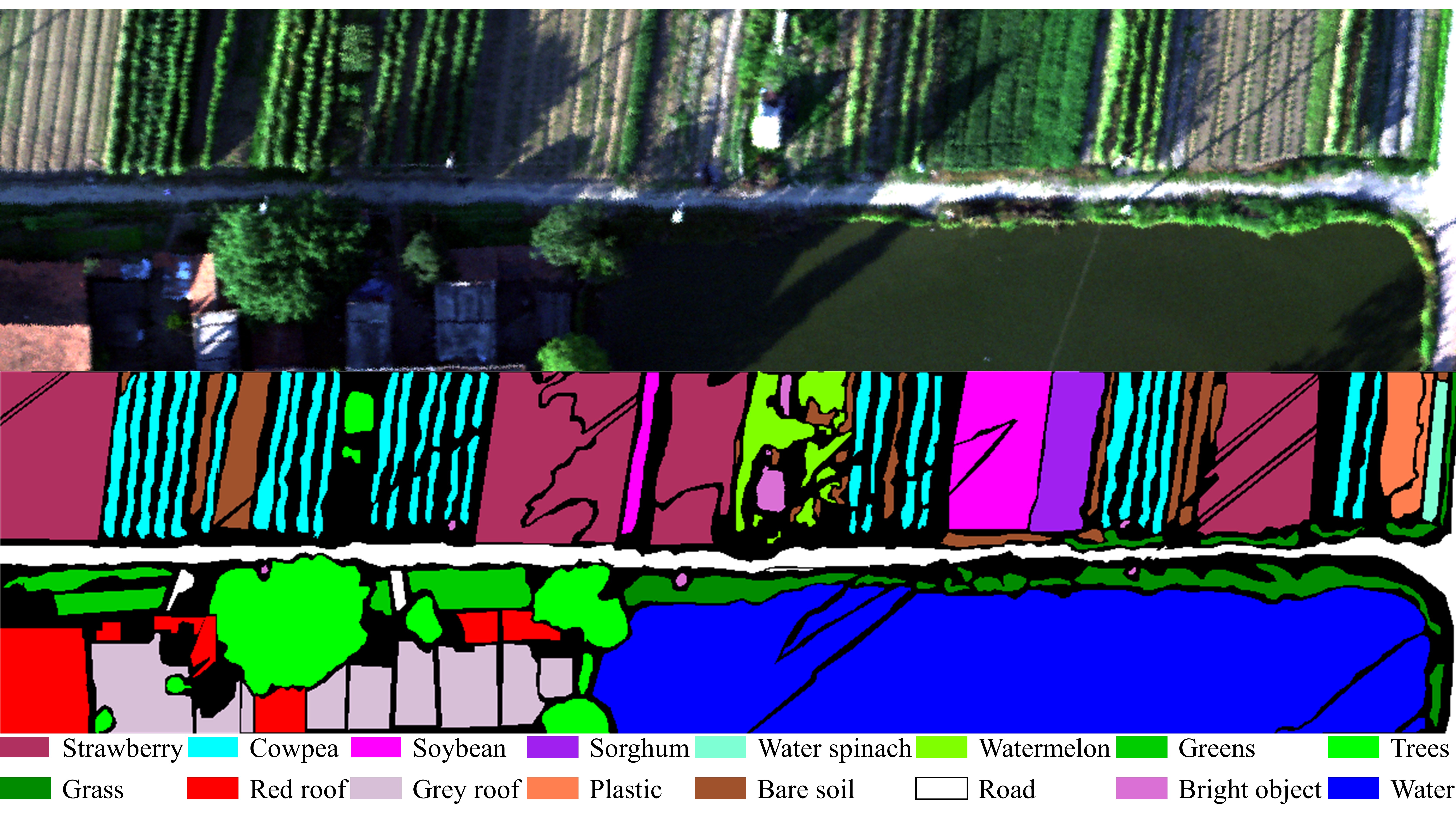}}
    \caption{Hyperspectral imagery with ground truth.
        (a) HongHu Dataset.
        (b) LongKou dataset.
        (c) HanChuan dataset.}
    \end{figure*}

\begin{table*}[!b]
    \tiny
    \caption{Details of the HSI Datasets (20 Classes in Total)}
    \label{HSI_dataset}
    \centering
    \resizebox{.98\textwidth}{!}{
    \begin{tabular}{ccccc}
    \toprule
    Dataset & Classes selected for one-class classification & \begin{tabular}[c]{@{}c@{}} Labeled samples \\ for each class \end{tabular} & \begin{tabular}[c]{@{}c@{}} Unlabeled samples \\ for each class\end{tabular} & \begin{tabular}[c]{@{}c@{}} Validation samples\\  for each class\end{tabular} \\
    \midrule
    HongHu   & \begin{tabular}[c]{@{}c@{}} cotton, rape, chinese cabbage, cabbage, tuber mustard, \\ brassica parachinensis, carrot, white radish (8 Classes)\end{tabular} & 100 & 4000 & 290878 \\
    LongKou  & corn, sesame, broad-leaf soybean, rice, water (5 Classes) & 100 & 4000 & 203642 \\
    HanChuan & \begin{tabular}[c]{@{}c@{}} strawberry, cowpea, soybean, water spinach,\\ watermelon, road, water (7 Classes)\end{tabular} & 100 & 4000 & 255930 \\
    \bottomrule
    \end{tabular}}
    \end{table*}

\subsubsection{Training Stable Strategies for the Proposed Risk Estimator}
In the aspect of the loss function, the sigmoid loss function is recommended in this paper to satisfy the assumption that the loss function is bounded.
Although cross-entropy loss is recommended as the loss function in the supervised classification task, this does not satisfy the assumption that the loss function is bounded and would result in instability in the proposed risk estimator.
Assuming that the estimated risk of the positive class is 0, then $R_{p}^{-}(f)$ will tend to infinity if cross-entropy loss is used.
In order to reduce the estimated risk of the negative classes, $R_{u}^{-}(f)$ will also tend to infinity, which will cause the network to identify all the training samples as positive.
We propose the use of the cross-entropy loss function to warm up the network, to avoid poor initial model parameters and obtain better initial model parameters, and then the sigmoid loss function is used to train the network.
In this study, the network was warmed up for 20 epochs.

Another potential source of instability comes from the dynamic modulating factor, which can become a very small number as $p_{i}$ approaches 1, leading to NaN in the gradient calculations, so probability clamping is used to limit the maximum of $p_{i}$ to 0.999.

\subsection{FreeOCNet in HOneCls}
\emph{FreeOCNet} is designed to extract global spectral-spatial features to verify the proposed risk estimator, which was inspired by recent work~\cite{9007624}.
This unified FCN is depicted in~Fig.~\ref{onecls}.

The basic module of the encoder is a spectral-spatial attention (SSA)-convolutional layer (Conv $3{\times}3$) with a group normalization (GN)- rectified linear unit (ReLU).
The SSA refines the features by weighting each pixel adaptively, which differs from spectral attention~\cite{9007624}, where weighting is conducted only in the spectral dimension.
The implementation of SSA is similar to that of a convolutional block attention module (CBAM)~\cite{10.1007/978-3-030-01234-2_1}.
The combination of a Conv $3{\times}3$-ReLU with stride 2 replaces the $3{\times}3$ pooling layer to reduce the spatial size of the feature maps, to align the center of the receptive field and its projected location.
A stem is placed at the front of the network to compress the data with different spectral channels into uniform channels.

A lightweight decoder with a fixed number of channels (128) is used in this FCN.
The decoder consists of an alternately stacked $3{\times}3$ convolutional layer and an upsampling layer with scale 2.

We use the fusion of high-level features and low-level features to obtain better semantic information and maintain better spatial details.
Differing from the fusion approach of UNet, the approach of point-wise addition between features with the same spatial size is adopted to fuse the high-level and low-level features, which makes the optimization easier by residual connection~\cite{7780459}.
The $1{\times}1$ convolution is adopted to adjust the number of channels in the low-level feature maps to match the channel requirement of the lightweight decoder.

\section{Experimental Results and Analysis}\label{sec:results}
\subsection{Experimental Settings}

\subsubsection{Dataset}
Three benchmark aerial hyperspectral images were used in the experiments, i.e., the WHU-Hi-HongHu dataset (Fig.~\ref{HongHu_dataset}), the WHU-Hi-LongKou dataset (Fig.~\ref{LongKou_dataset}), and the WHU-Hi-HanChuan dataset (Fig.~\ref{HanChuan_dataset})~\cite{ZHONG2020112012}.
The classes in these hyperspectral images are similar in texture and spectra.
Therefore, one-class classification on these three datasets is a huge challenge.
Please refer to~\cite{ZHONG2020112012} for more detailed description of datasets.

One hundred positive samples and unlabeled samples, which were 40 times the number of positive samples, were randomly selected from the images to participate in the training.
In addition to some inaccurate annotations, 20 kinds of ground objects were selected for the one-class classification.
More detailed information about the HSI datasets is provided in Table~\ref{HSI_dataset}.
In order to avoid the impact of inaccurate estimation of the class prior on the results of the PU learning, we adopted the approximate real class prior estimated from the ground truth.

\subsubsection{Training Details}
In order to demonstrate the robustness of the proposed method, all the ground objects were classified with the same hyperparameters in the proposed \emph{HOneCls}.
To make a fair comparison in the deep learning-based methods, we also aligned the hyperparameters of all the deep learning-based methods.
All deep learning-based methods were trained for 1000 epochs using a stochastic gradient descent optimizer.
In addition, the weight decay was set to 0.0001 and the momentum was set to 0.9 for all the experiments, with no modification.
${\alpha}_{p}=0.3$ and ${\gamma}=0.1$ were the default settings, and experiments were conducted to determine these two parameters.

\subsubsection{Metrics}
In this paper, the methods are mainly evaluated using the F1-score ($\times$100), precision and recall are shown in Appendix B as supplement.
Precision represents how many of the positive samples predicted by the classifier are truly positive samples, and recall is how many truly positive samples are correctly predicted by the classifier.
As the harmonic average of the precision and recall, the F1 score is suitable for the scenarios of one-class classification:
\begin{equation}
\nonumber
F1-score = \frac{2 \times Precision \times Recall}{Precision + Recall}.
\end{equation} 
What's more, the average F1-score of different ground objects was calculated to test the robustness of the one-class classification algorithm on different ground objects.
Since different datasets were shot under different conditions, the average F1-score was calculated for different ground objects on the same dataset.
All the experiments were repeated five times.

\subsubsection{Methods}
As a control experiment, unlabeled samples were treated as negative data trained by binary cross-entropy (BCE) loss, to show that it is inappropriate to simply treat unlabeled data as negative data.

We compared the proposed method with the other one-class classification methods in the remote sensing community: one-class support vector machine (OCSVM)~\cite{NIPS1999_8725fb77}, the positive and unlabeled learning algorithm (PUL)~\cite{LU2021112584}, the positive and background learning algorithm (PBL)~\cite{9201373}, and biased support vector machine (BSVM)~\cite{PIIROINEN2018119}.
The basic classifiers in PUL and PBL are replaced with deep learning based classifier and are denoted as DeepPUL and DeepPBL.
The proposed method was compared with deep unbiased risk estimation based methods, DOCC~\cite{LEI2021102598} and absNegative~\cite{ZHAO2022328}, to demonstrate the importance of considering the problem of distribution imbalance in HSI.
\emph{FreeOCNet} was used as the global spectral-spatial feature extractor for BCE, DeepPUL, DeepPBL, DOCC, absNegative, and \emph{HOneCls}, the configuration details of the FreeOCNet is listed in Table ~\ref{FreeOCNet_config}.
.

\begin{table}[!b]
    \caption{Configuration details of the FreeOCNet}
    \label{FreeOCNet_config}
    \centering
    \resizebox{.98\columnwidth}{!}{
    \begin{tabular}{ccccc}
    \toprule
    Stem                                                            & Encoder                                                                                                                            & Lateral     & Decoder                                                                      \\ \midrule
    \begin{tabular}[c]{@{}c@{}}Conv3x3:64;\\ Groups:16\end{tabular} & \begin{tabular}[c]{@{}c@{}}\#1-\#4 Conv3x3:\\ 128,192,256,320;\\ Groups:16;\\ \#1-\#3 Conv3x3,stride 2:\\ 192,256,320\end{tabular} & Conv1x1:128 & \begin{tabular}[c]{@{}c@{}}Conv3x3:128;\\ Nearest interpolation\end{tabular} \\ \bottomrule
\end{tabular}}
\end{table}

The HSI target detection methods, including ECEM~\cite{rs11111310}, DSM~\cite{9884362} and G2LHTD~\cite{9968036}, were also compared. OTSU is used to select the threshold for HSI target detection methods.

\begin{figure*}[!t]
\centering
\includegraphics[width=0.98\textwidth]{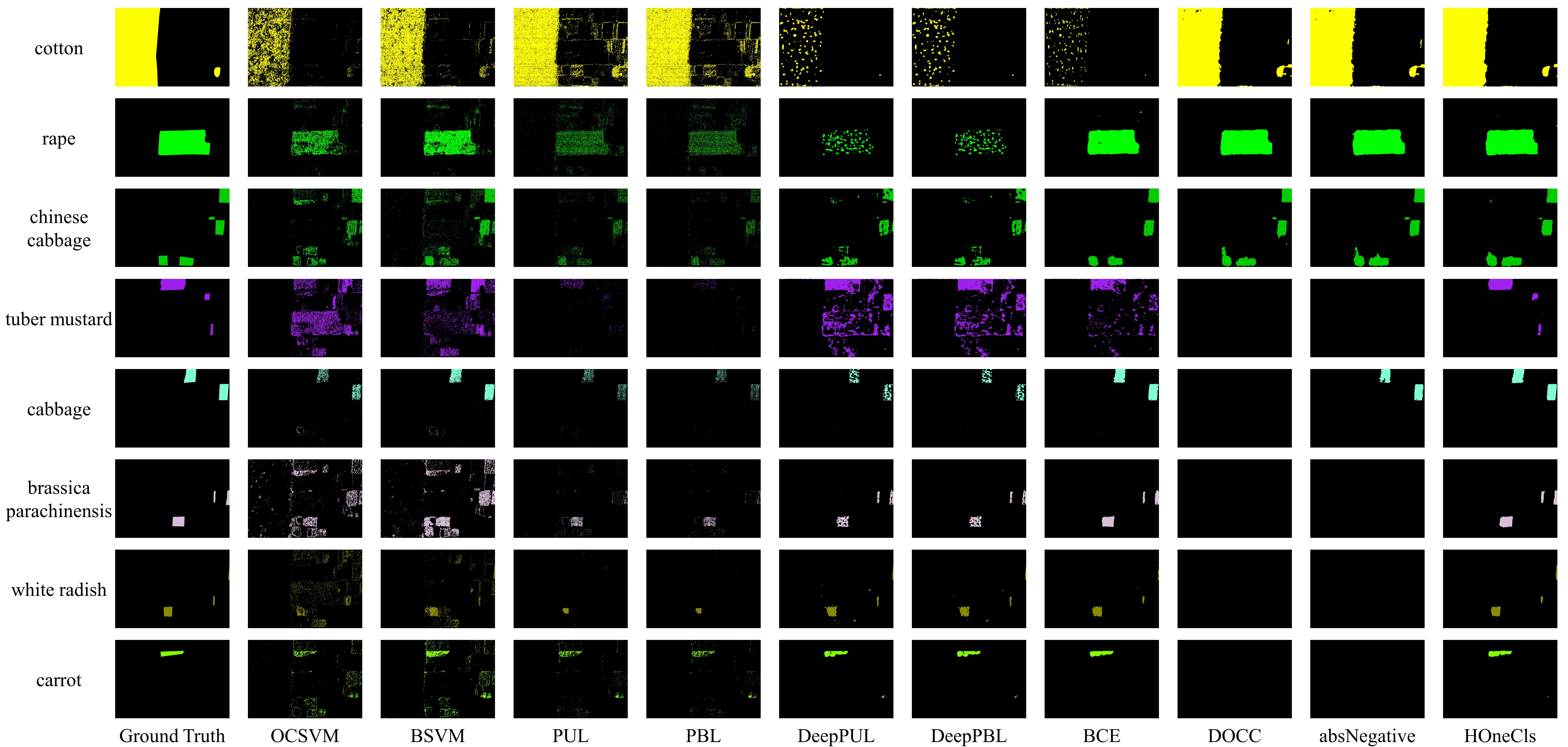}
\caption{Distribution maps for the HongHu dataset.}
\label{HongHu_result}
\end{figure*}

\begin{table*}[!b]
    \tiny
    \caption{The F1 scores for the HongHu dataset. The mean and standard deviation are reported.}
    \label{HongHu_f1}
    \centering
    \resizebox{.98\textwidth}{!}{
        \begin{tabular}{cccccccccccc}
            \toprule
            Class                  & Class prior & OCSVM       & BSVM        & PUL         & PBL         & DeepPUL      & DeepPBL      & BCE          & DOCC         & absNegative  & HOneCls              \\ \midrule
            cotton                 & 0.3769      & 60.34(0.00) & 87.49(0.00) & 91.50(0.47) & 90.62(0.62) & 16.70(2.89)  & 16.62(2.87)  & 10.02(0.95)  & 99.19(0.51)  & 99.31(0.18)  & \textbf{99.33(0.26)} \\
            rape                   & 0.1317      & 59.96(0.00) & 83.83(0.00) & 44.95(0.84) & 41.08(1.08) & 60.91(33.16) & 60.76(33.41) & 55.83(37.80) & 94.70(8.78)  & 98.57(0.08)  & \textbf{98.59(0.34)} \\
            chinese cabbage        & 0.0544      & 55.89(0.00) & 55.89(0.00) & 43.20(3.67) & 37.39(2.40) & 79.32(14.35) & 79.62(14.23) & 79.77(13.31) & 34.11(46.76) & 53.66(49.00) & \textbf{90.59(1.26)} \\
            tuber mustard          & 0.0367      & 17.41(0.00) & 36.40(0.00) & 24.18(4.93) & 11.07(4.35) & 59.67(10.48) & 59.62(8.00)  & 58.88(2.48)  & 0.00(0.00)   & 0.00(0.00)   & \textbf{95.52(0.46)} \\
            cabbage                & 0.0319      & 62.71(0.00) & 92.59(0.00) & 44.51(2.98) & 34.12(5.27) & 82.55(18.93) & 81.75(19.41) & 79.42(20.44) & 0.00(0.00)   & 18.66(41.72) & \textbf{98.39(0.40)} \\
            brassica parachinensis & 0.0194      & 29.16(0.00) & 35.77(0.00) & 54.03(0.54) & 48.55(0.99) & 88.55(5.09)  & 87.86(5.58)  & 87.06(5.98)  & 0.00(0.00)   & 0.00(0.00)   & \textbf{94.56(1.11)} \\
            white radish           & 0.0119      & 23.07(0.00) & 52.76(0.00) & 40.92(0.48) & 38.36(0.68) & 87.66(4.08)  & 86.39(5.17)  & 86.53(5.13)  & 0.00(0.00)   & 0.00(0.00)   & \textbf{94.30(0.32)} \\
            carrot                 & 0.0102      & 32.60(0.00) & 44.59(0.00) & 49.61(0.23) & 47.10(0.20) & 77.97(18.87) & 82.06(13.20) & 81.21(14.38) & 0.00(0.00)   & 0.00(0.00)   & \textbf{93.58(0.82)} \\ \midrule
            Average F1-score       &             & 42.64(0.00) & 62.30(0.00) & 49.11(1.77) & 43.53(1.95) & 69.17(13.48) & 69.33(12.73) & 67.34(12.56) & 28.50(7.01)  & 33.77(11.37) & \textbf{95.61(0.62)} \\ \bottomrule
\end{tabular}}
\end{table*}

\begin{table*}[!b]
    \tiny
    \caption{The F1-scores for the LongKou dataset. The mean and standard deviation are reported.}
    \label{LongKou_f1}
    \centering
    \resizebox{.98\textwidth}{!}{
        \begin{tabular}{cccccccccccc}
            \toprule
            Class              & Class prior & OCSVM       & BSVM        & PUL         & PBL         & DeepPUL      & DeepPBL      & BCE          & DOCC                 & absNegative          & HOneCls              \\ \midrule
water              & 0.3048      & 68.80(0.00) & 88.35(0.00) & 78.44(4.48) & 78.14(5.84) & 80.36(12.49) & 80.24(12.55) & 63.64(20.15) & 98.84(0.17)          & 96.88(0.13)          & \textbf{98.89(0.16)} \\
broad-leaf soybean & 0.2873      & 63.58(0.00) & 78.12(0.00) & 80.22(1.53) & 79.72(1.27) & 20.20(1.24)  & 20.11(1.23)  & 14.53(0.71)  & 90.67(0.59)          & \textbf{95.88(0.63)} & 92.60(3.11)          \\
corn               & 0.1569      & 67.99(0.00) & 93.97(0.00) & 79.93(1.28) & 77.64(1.69) & 33.17(1.53)  & 32.94(1.50)  & 26.05(1.89)  & \textbf{99.66(0.04)} & 99.63(0.05)          & 99.58(0.05)          \\
rice               & 0.0539      & 65.24(0.00) & 94.96(0.00) & 78.43(2.37) & 67.83(3.51) & 61.47(6.35)  & 60.68(6.07)  & 55.05(4.60)  & 73.23(42.31)         & 79.65(44.53)         & \textbf{99.92(0.02)} \\
sesame             & 0.0138      & 24.34(0.00) & 68.95(0.00) & 65.67(0.92) & 59.80(2.61) & 95.81(2.37)  & 95.13(3.70)  & 95.01(3.90)  & 0.00(0.00)           & 0.00(0.00)           & \textbf{99.39(0.08)} \\ \midrule
Average F1-score   &             & 57.99(0.00) & 84.87(0.00) & 76.54(2.12) & 72.63(2.98) & 58.20(4.80)  & 57.82(5.01)  & 50.85(6.25)  & 72.48(8.62)          & 74.41(9.07)          & \textbf{98.07(0.68)} \\ \bottomrule
\end{tabular}}
\end{table*}

\begin{table*}[!b]
    \tiny
    \caption{The F1-scores for the HanChuan dataset. The mean and standard deviation are reported.}
    \label{HanChuan_f1}
    \centering
    \resizebox{.98\textwidth}{!}{
        \begin{tabular}{cccccccccccc}
            \toprule
            Class            & Class prior & OCSVM       & BSVM        & PUL         & PBL         & DeepPUL      & DeepPBL      & BCE          & DOCC        & absNegative          & HOneCls              \\ \midrule
water            & 0.2045      & 60.81(0.00) & 94.38(0.00) & 56.64(4.53) & 57.02(4.81) & 90.21(5.58)  & 90.32(5.47)  & 81.47(29.73) & 97.91(0.29) & \textbf{97.93(0.15)} & 97.82(0.12)          \\
strawberry       & 0.1213      & 67.86(0.00) & 80.43(0.00) & 7.32(3.51)  & 4.18(1.61)  & 77.33(24.07) & 77.30(24.28) & 72.40(27.35) & 87.64(5.93) & 90.05(5.65)          & \textbf{93.68(0.49)} \\
cowpea           & 0.0617      & 34.46(0.00) & 45.38(0.00) & 4.40(0.91)  & 4.44(0.33)  & 43.20(0.93)  & 42.62(0.91)  & 37.49(0.63)  & 0.00(0.00)  & 0.00(0.00)           & \textbf{85.66(3.01)} \\
road             & 0.0503      & 37.46(0.00) & 65.81(0.00) & 2.99(3.77)  & 2.86(4.09)  & 76.87(12.38) & 76.27(12.59) & 70.69(13.30) & 0.00(0.00)  & 0.00(0.00)           & \textbf{91.56(1.14)} \\
soybean          & 0.0279      & 49.85(0.00) & 46.27(0.00) & 34.86(1.59) & 25.67(2.49) & 68.64(19.64) & 68.18(18.42) & 67.05(15.24) & 0.00(0.00)  & 0.00(0.00)           & \textbf{96.92(3.64)} \\
watermelon       & 0.0123      & 13.99(0.00) & 17.80(0.00) & 6.83(1.96)  & 2.62(1.57)  & 83.33(3.85)  & 82.39(4.10)  & 82.42(4.09)  & 0.00(0.00)  & 0.00(0.00)           & \textbf{91.17(1.05)} \\
water spinach    & 0.0033      & 21.65(0.00) & 19.47(0.00) & 33.25(0.28) & 11.96(1.46) & 96.72(1.48)  & 96.29(1.85)  & 96.67(1.56)  & 0.00(0.00)  & 0.00(0.00)           & \textbf{98.63(1.28)} \\ \midrule
Average F1-score &             & 40.87(0.00) & 52.79(0.00) & 20.90(2.36) & 15.54(2.34) & 76.62(9.71)  & 76.19(9.66)  & 72.59(13.13) & 26.51(0.89) & 26.85(0.83)          & \textbf{93.63(1.53)} \\ \bottomrule
\end{tabular}}
\end{table*}

\subsection{Experiment 1: HongHu Dataset}
The HongHu HSI has 270 channels from 400 to 1000 nm, and the spatial resolution of this imagery is about 0.043 m.
The HongHu data used in this study was selected from the original HongHu imagery to remove most of the pixels without the labels.
The selected hyperspectral image was 678$\times$465 pixels in size , and eight unambiguous land-cover types were selected for the one-class classification.

The distribution maps are shown in Fig.~\ref{HongHu_result}, and the F1-scores are listed in Table~\ref{HongHu_f1}.
The precision and recall are shown in Appendix B as supplement.
From the distribution maps, \emph{HOneCls} accurately identifies all the land-cover types.
The \emph{HOneCls} framework obtains the best F1-score in all the selected target objects, and the average F1-score (95.61) of \emph{HOneCls} is significantly higher than that of the other classical one-class classification methods, which indicates that \emph{HOneCls} shows good robustness on different objects.
The risk estimation based methods (DOCC, absNegative, \emph{HOneCls})  has better performance on HSI with distribution overlap.
However, DOCC and absNegative do not recognize the positive class for the distribution imbalanced ground objects, i.e., under-fitting of the positive class occurrs, and the peoposed \emph{HOneCls} has the ability to mapping ground objects accurately in the case of distribution imbalance.

Another interesting finding is that BCE outperforms existing one-class classification methods in the remote sensing community for some low class prior targets, such as tuber mustard, brassica parachinensis, white radish and carrot.
The low target probability means that there are fewer positive samples in the unlabeled data, when the BCE will be less affected by mislabeled negative training samples, however, the BCE performs worse in one-class classification of ground objects with a large proportion of classes, such as cotton and rape.
The method proposed in this paper significantly improves the average F1-score by 26.28 compared to the second-best method.

\begin{figure*}[!t]
    \centering
    \includegraphics[width=0.98\textwidth]{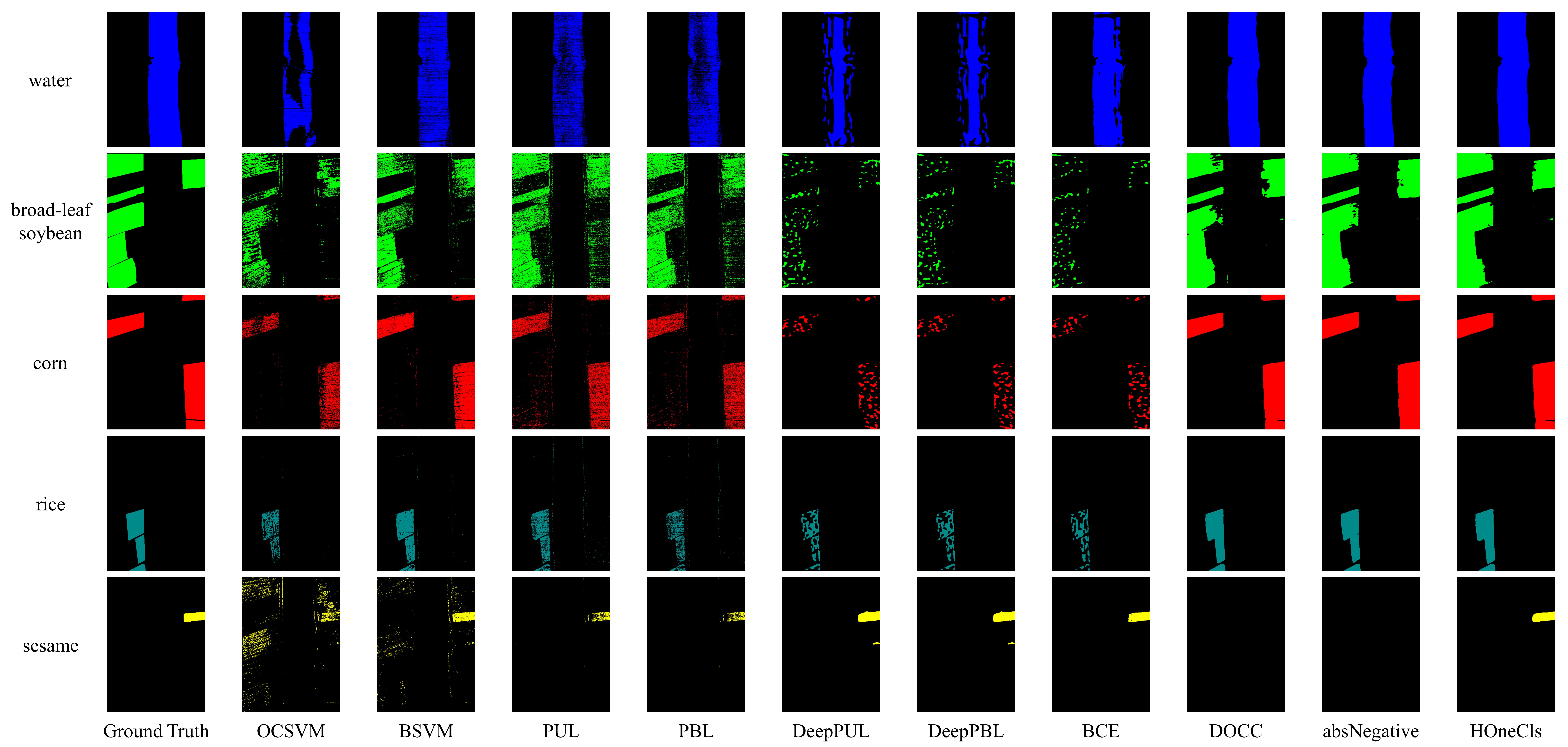}
    \caption{Distribution maps for the LongKou dataset.}
    \label{LongKou_result}
\end{figure*}

\begin{figure*}[!t]
    \centering
    \includegraphics[width=0.98\textwidth]{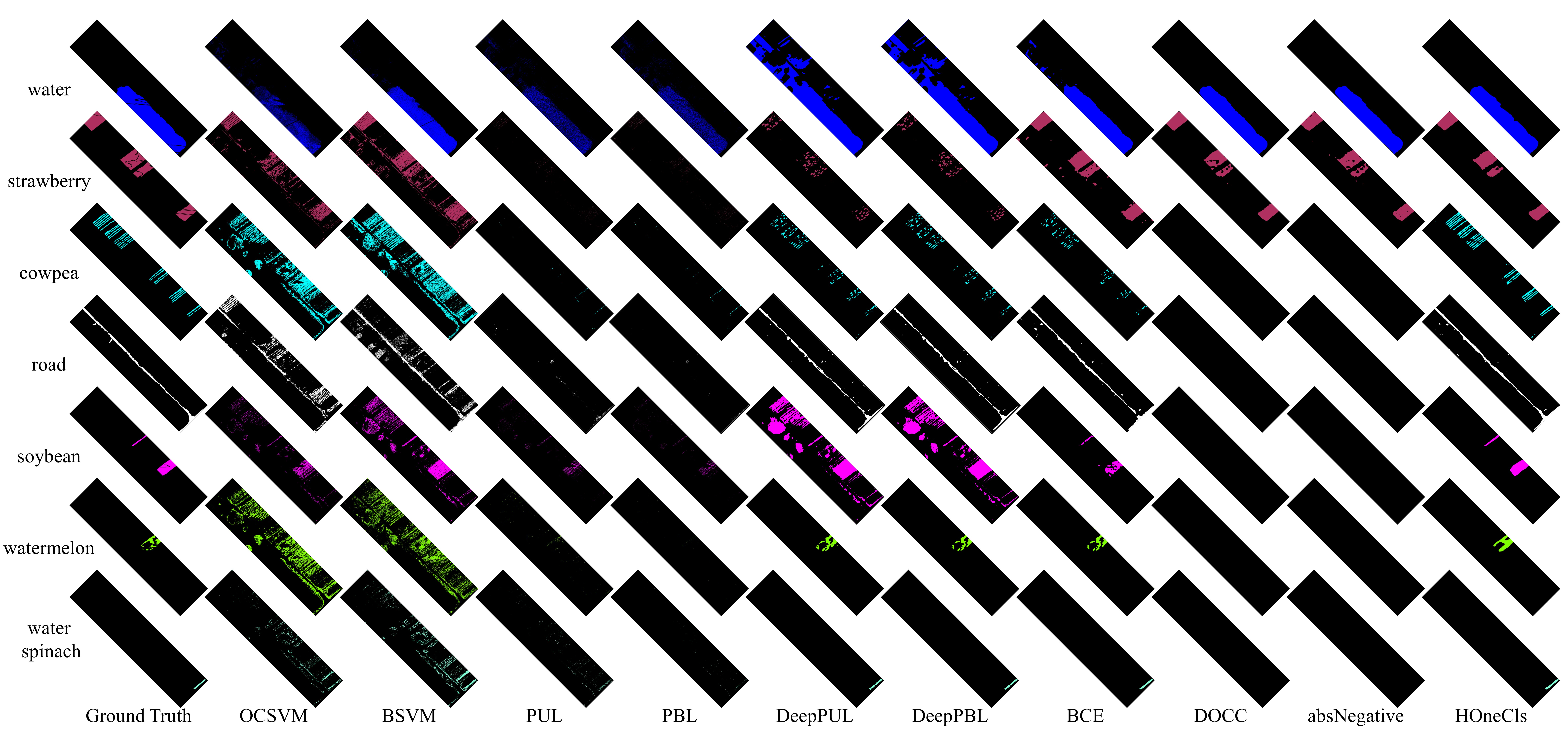}
    \caption{Distribution maps for the HanChuan dataset.}
    \label{HanChuan_result}
\end{figure*}

\subsection{Experiment 2: LongKou Dataset}

The LongKou HSI has 270 bands from 400 to 1000 nm, and the spatial resolution is about 0.463 m.
The image is of 550$\times$400 pixels in size.
To avoid the imprecise labels, five of the nine ground objects are selected for the one-class classification.

The detection maps for the LongKou dataset are shown in Fig.~\ref{LongKou_result}, and the F1-scores are listed in Table~\ref{LongKou_f1}.
The precision and recall are shown in Appendix B as supplement.
Similar to the results for the HongHu dataset, \emph{HOneCls} shows good robustness to the ground objects in the LongKou dataset.
From the distribution maps, it can be seen that the risk estimation based methods obtain better one-class classification results, but DOCC and absNegative do not recognize the positive class for the distribution imbalanced ground objects, i.e., the under-fitting of the positive class occurrs.
The \emph{HOneCls} framework obtains good F1-scores in all the land-cover types and \emph{HOneCls} accurately identifies all the land-cover types.
The average F1-score of \emph{HOneCls} is 98.07, and the second best average F1-score is BSVM with 84.87.
The results of BCE show that supervised binary HSI classification method is not suitable for the one-class classification task, and the results show obvious overfitting.

\subsection{Experiment 3: HanChuan Dataset}

The HanChuan imagery has 274 channels from 400 to 1000 nm, and the image is of 1217$\times$303 pixels in size.
The spatial resolution of the HanChuan imagery is 0.109 m.
In particular, the HanChuan dataset is overlaid with distinct shadows, which significantly increases the spectral variability.
Seven kinds of ground objects without obvious imprecise labeling were selected for detection.

The distribution maps for the HanChuan dataset (Fig.~\ref{HanChuan_result}) also show the improvement of the robustness of the \emph{One-Class Risk Estimator} for the ground targets in the HanChuan dataset.
\emph{HOneCls} obtains excellent one-class classification results, for both the distribution balanced and imbalanced data, for all the classes.
The F1-scores for the HanChuan dataset are listed in Table~\ref{HanChuan_f1}.
The precision and recall are shown in Appendix B as supplement.
The \emph{HOneCls} framework obtain a good F1-score in all the land-cover types.
The \emph{One-Class Risk Estimator} performs slightly worse than absNegative in the water class, but the proposed risk estimator shows a significant improvement in average F1-score.
The average F1-score of \emph{HOneCls} is 93.63, which is significantly higher than that of the second best method with an improved F1-score of 17.01.

\subsection{Comparison with Multi-class Classification Methods}

Experiments were conducted to analyze the performance of HSI multi-class classification methods (FPGA~\cite{9007624}, DBDA-MISH~\cite{rs12030582} and HPDM-SPRN~\cite{9454961}) and one-class classification methods.
In the multi-class classification methods, unlabeled samples were treated as negative.
The classes "cotton" and "rape" in HongHu dataset were selected as examples, and the results are shown in Table~\ref{MCC_f1}.
From Table~\ref{MCC_f1}, it can be inferred that using multi-class classification methods in one-class scenarios is inappropriate.

\subsection{Comparison with HSI Target Detection Methods}

Experiments were conducted to analyze the performance of HSI target detection methods and one-class classification methods.
The classes "tuber mustard", "brassica parachinensis" and "carrot" in HongHu dataset were selected as examples.
The F1-score is shown in Table~\ref{HTD_f1}, the Area Under the Curve (AUC) is shown in Table~\ref{HTD_auc}.
The proposed \emph{HOneCls} achieved the best F1-score and AUC score.

\subsection{Ablation Experiments for One-Class Risk Estimation}

Ablation experiments were conducted to analyze conformance based negative risk estimation strategy and positive representation enhancement strategy in Table~\ref{tab:one_class_risk_ablation}.
Three conclusions can be obtained from Table~\ref{tab:one_class_risk_ablation}:
(1) The performance of supervised binary classification methods is limited in HSI one-class classification scenarios.
Compared with BCE and Focal Loss, the proposed \emph{HOneCls} obtain the best F1-score.
Although the positive representation enhancement strategy is inspired from Focal Loss, the \emph{HOneCls} is more robust to the noisy labels (unlabeled data) than Focal Loss.
(2) The High-quality negative risk estimation is more conducive to deep learning-based one-class classification.
Compared with $\widehat{R}_{n}^-(f)$, the $\widehat{R}_{oc-n}^-(f)$ obtain better F1-score due to the adaptive ``flood level''.
(3) The positive representation enhancement strategy is a simply but effective solution for overcoming the problem of distribution imbalance.
In the experiments of tuber mustard (class probability: 0.0367), the F1-score is 0.00 and 95.52 with and without positive representation enhancement strategy, respectively.

\begin{table}[!b]
    \tiny
    \caption{The F1 scores of both the multi-class and one-class classification methods. The mean and standard deviation are reported.}
    \label{MCC_f1}
    \centering
    \resizebox{.95\columnwidth}{!}{
    \begin{tabular}{ccccc}
    \toprule
    Class  & FPGA         & DBDA-MISH    & HPDM-SPRN   & HOneCls              \\ \midrule
cotton & 10.15(0.97)  & 0.00(0.00)   & 1.48(1.58)  & \textbf{99.33(0.26)} \\
rape   & 82.38(24.34) & 14.54(31.91) & 9.15(15.96) & \textbf{98.59(0.34)} \\ \bottomrule
    \end{tabular}}
    \end{table}

\begin{table}[!b]
\tiny
\caption{The F1 scores of both the target detection and one-class classification methods. The mean and standard deviation are reported.}
\label{HTD_f1}
\centering
\resizebox{.95\columnwidth}{!}{
\begin{tabular}{ccccc}
\toprule
Class                  & ECEM        & DSM         & G2LHTD       & HOneCls              \\ \midrule
tuber mustard          & 24.05(0.00) & 21.33(0.12) & 29.25(0.00)  & \textbf{95.52(0.46)} \\
brassica parachinensis & 17.68(0.00) & 17.24(0.87) & 10.92(0.00)  & \textbf{94.56(1.11)} \\
carrot                 & 9.28(0.00)  & 3.20(0.06)  & 17.20(0.00)  & \textbf{93.58(0.82)} \\ \bottomrule
\end{tabular}}
\end{table}

\begin{table}[!b]
\tiny
\caption{The AUC of both the target detection and one-class classification methods. The mean and standard deviation are reported.}
\label{HTD_auc}
\centering
\resizebox{.95\columnwidth}{!}{
\begin{tabular}{ccccc}
\toprule
Class                  & ECEM        & DSM         & G2LHTD      & HOneCls              \\ \midrule
tuber mustard          & 80.89(0.00) & 84.40(0.07) & 66.99(0.00) & \textbf{99.68(0.12)} \\
brassica parachinensis & 91.46(0.00) & 88.78(0.06) & 65.43(0.00) & \textbf{99.98(0.01)} \\
carrot                 & 65.55(0.00) & 77.75(0.08) & 81.03(0.00) & \textbf{99.98(0.01)} \\ \bottomrule
\end{tabular}}
\end{table}

\begin{table}[!b]
    \centering
    \caption{Ablation Analysis of One-Class Risk Estimator}
    \label{tab:one_class_risk_ablation}
    \resizebox{.98\columnwidth}{!}{  
        \begin{tabular}{cccccc}
            \toprule
            Class                                                                                         & Methods                  & \begin{tabular}[c]{@{}c@{}}Positive representation\\  enhancement strategy\end{tabular} & $\widehat{R}_{n}^-(f)$ & $\widehat{R}_{oc-n}^-(f)$ & F1-score       \\ \midrule
            \multirow{6}{*}{\begin{tabular}[c]{@{}c@{}}rape\\ Probability:0.1317\end{tabular}}            & BCE                      & -                                                                                       & -                      & -                         & 55.83          \\
                                                                                                          & Focal Loss               & -                                                                                       & -                      & -                         & 24.53          \\ \cmidrule{2-6}
                                                                                                          & \multirow{4}{*}{HOneCls} &                                                                                         & \checkmark              &                           & 94.25          \\
                                                                                                          &                          &                                                                                         &                        & \checkmark                 & 98.57          \\
                                                                                                          &                          & \checkmark                                                                               & \checkmark              &                           & 96.28          \\
                                                                                                          &                          & \checkmark                                                                               &                        & \checkmark                 & \textbf{98.59} \\ \midrule
            \multirow{6}{*}{\begin{tabular}[c]{@{}c@{}}tuber\\ mustard\\ Probability:0.0367\end{tabular}} & BCE                      & -                                                                                       & -                      & -                         & 58.88          \\
                                                                                                          & Focal Loss               & -                                                                                       & -                      & -                         & 61.23          \\ \cmidrule{2-6}
                                                                                                          & \multirow{4}{*}{HOneCls} &                                                                                         & \checkmark              &                           & 0.00           \\
                                                                                                          &                          &                                                                                         &                        & \checkmark                 & 0.00           \\
                                                                                                          &                          & \checkmark                                                                               & \checkmark              &                           & 93.29          \\
                                                                                                          &                          & \checkmark                                                                               &                        & \checkmark                 & \textbf{95.52} \\ \bottomrule
            \end{tabular}}
    \end{table}

\subsection{Analysis of ${\alpha}_{p}$ and ${\gamma}$}
Ablation experiments were carried out to analyze ${\alpha}_{p}$ and ${\gamma}$.
The purpose of this paper is to propose a one-class classification algorithm that can be applied to most of the ground objects, we therefore chose to select a set of parameters that performed well in most of the ground objects by reporting the average F1-score of all class of objects on the HongHu dataset~(Table~\ref{FocalOneClassRiskEstimator}).
The experiment was repeated five times, and the average F1-score and the average standard deviation are reported.

\begin{table*}[!b]
    \centering
    \caption{Ablation experiments in One-Class Risk Estimation for ${\alpha}_{p}$ and ${\gamma}$ on the HongHu dataset. The mean and standard deviation are reported.}
    \label{FocalOneClassRiskEstimator}
    \subfloat[Varying $\alpha$ ($\gamma$ = 0)]{
    \begin{tabular}{ccccccccccc}
    \toprule
    $\alpha$ & $\gamma$ & Average F1       \\
    \midrule
    0.1      & 0 & 74.09 (21.12)         \\
    0.2      & 0 & 93.18 (6.12)          \\
    0.3      & 0 & \textbf{93.64 (5.13)} \\
    0.4      & 0 & 89.93 (12.38)         \\
    0.5      & 0 & 93.09 (5.63)          \\
    0.6      & 0 & 91.85 (7.02)          \\
    \bottomrule
    \end{tabular}}
    \hspace{0.5mm}
    \subfloat[Varying $\gamma$ ($\alpha$ = 0.2)]{
    \begin{tabular}{ccccccccccc}
    \toprule
    $\alpha$ & $\gamma$ & Average F1            \\
    \midrule
    0.2      & 0.0      & 93.18 (6.12)          \\
    0.2      & 0.1      & 90.95 (7.06)          \\
    0.2      & 0.3      & 94.35 (3.86)          \\
    0.2      & 0.5      & \textbf{95.53 (0.66)} \\
    0.2      & 1.0      & 95.16 (0.79)          \\
    0.2      & 3.0      & 94.45 (0.94)          \\
    \bottomrule
    \end{tabular}}
    \hspace{0.5mm}
    \subfloat[Varying $\gamma$ ($\alpha$ = 0.3)]{
    \begin{tabular}{ccccccccccc}
    \toprule
    $\alpha$ & $\gamma$ & Average F1            \\
    \midrule
    0.3      & 0.0      & 93.64 (5.13)          \\
    0.3      & 0.1      & \textbf{95.61 (0.62)} \\
    0.3      & 0.3      & 95.53 (0.95)          \\
    0.3      & 0.5      & 94.21 (3.78)          \\
    0.3      & 1.0      & 94.89 (1.12)          \\
    0.3      & 3.0      & 92.99 (4.28)          \\
    \bottomrule
    \end{tabular}}
    \hspace{0.5mm}
    \subfloat[Varying $\gamma$ ($\alpha$ = 0.4)]{
    \begin{tabular}{ccccccccccc}
    \toprule
    $\alpha$ & $\gamma$ & Average F1            \\
    \midrule
    0.4      & 0.0      & 89.93 (12.38)         \\
    0.4      & 0.1      & 94.00 (3.04)          \\
    0.4      & 0.3      & 94.40 (3.72)          \\
    0.4      & 0.5      & \textbf{95.49 (0.81)} \\
    0.4      & 1.0      & 95.26 (0.98)          \\
    0.4      & 3.0      & 93.22 (3.80)          \\
    \bottomrule
    \end{tabular}}
    \hspace{0.5mm}
    \subfloat[Varying $\gamma$ ($\alpha$ = 0.5)]{
    \begin{tabular}{ccccccccccc}
    \toprule
    $\alpha$ & $\gamma$ & Average F1            \\
    \midrule
    0.5      & 0.0      & 93.09 (5.63)          \\
    0.5      & 0.1      & 92.12 (6.19)          \\
    0.5      & 0.3      & 93.82 (3.69)          \\
    0.5      & 0.5      & 93.58 (5.19)          \\
    0.5      & 1.0      & \textbf{94.42 (3.04)} \\
    0.5      & 3.0      & 92.14 (4.69)          \\
    \bottomrule
    \end{tabular}}
    \end{table*}

\begin{figure*}[t]
\centering
\includegraphics[width=0.98\textwidth]{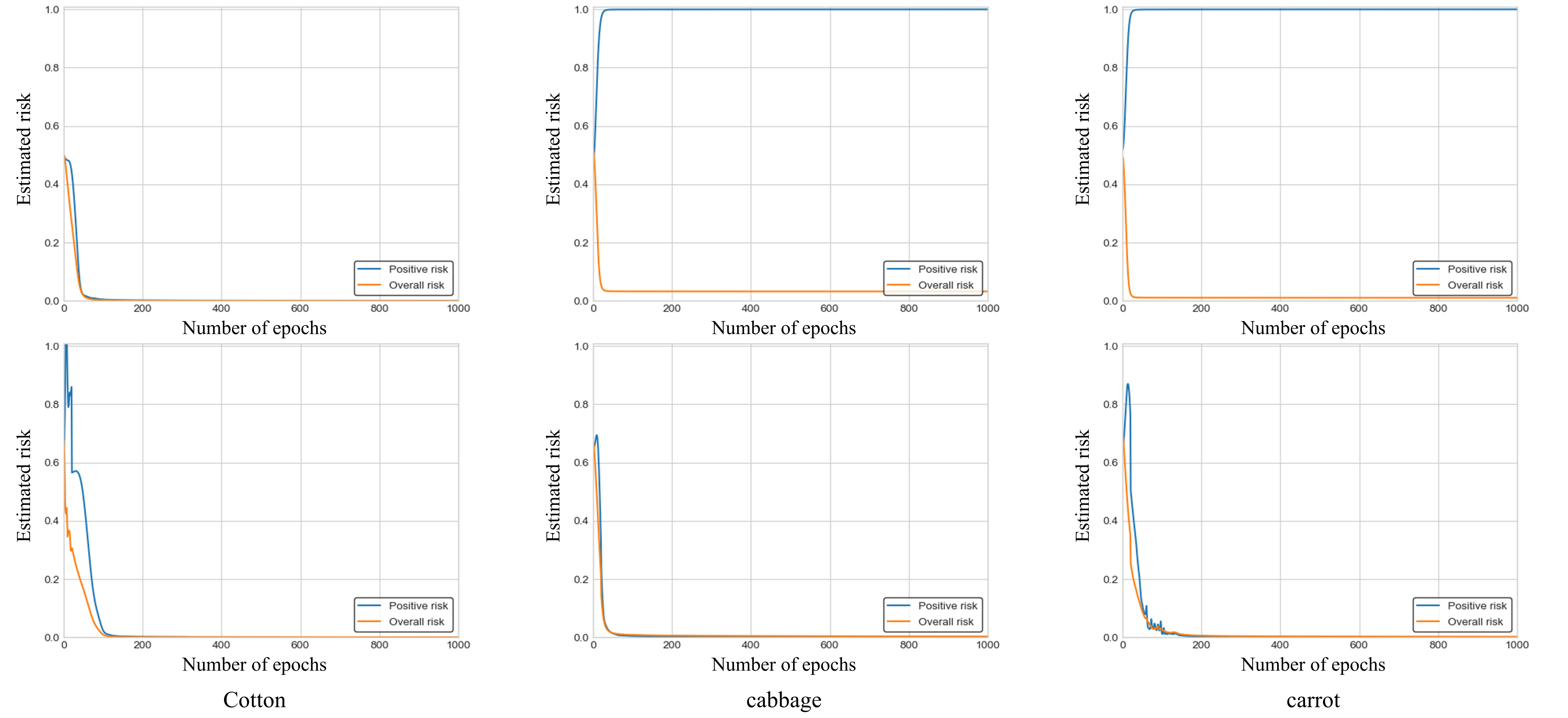}
\caption{Risk in the training stage.
\textbf{First row:} When the class prior is small, the risk of the positive class does not decrease without considering the problem of distribution imbalance.The class prior of the cotton class is larger, and the risk is normally reduced.
\textbf{Second row:} After considering the problem of distribution imbalance, the risk of the positive class decreases normally.}
\label{Risk_Change}
\end{figure*}

\begin{figure*}[!t]
    \centering
    \includegraphics[width=0.98\textwidth]{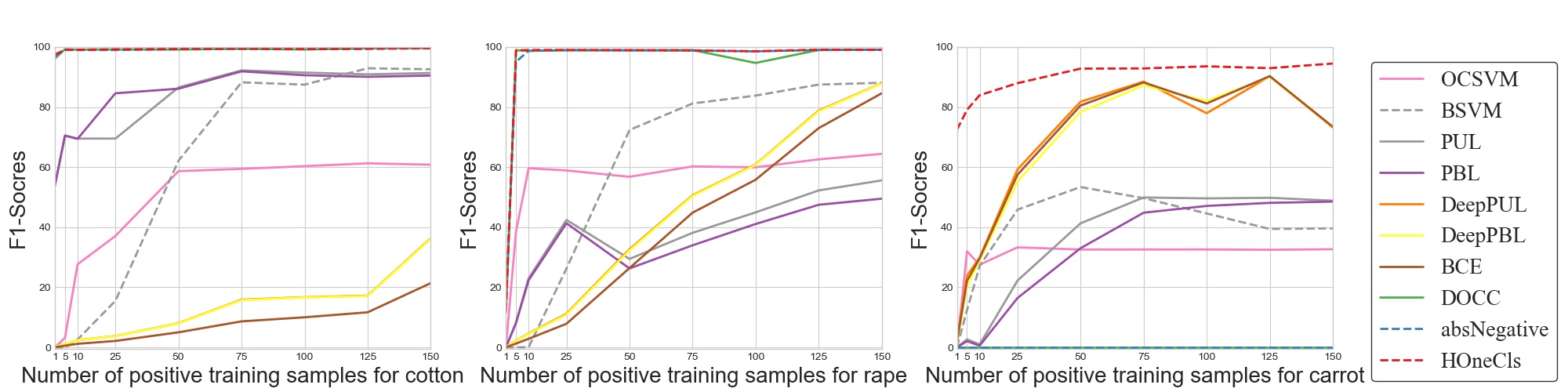}
    \caption{Performance versus the number of positive training samples.}
    \label{training_samples_results}
    \end{figure*}

\begin{figure*}[!t]
    \centering
    \includegraphics[width=0.98\textwidth]{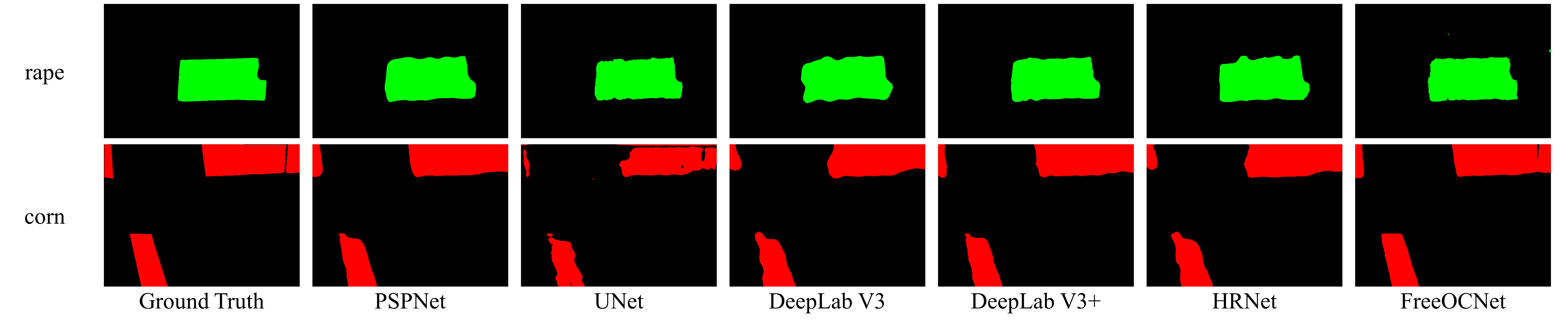}
    \caption{The distribution maps for different FCNs.
    The first row shows the results for rape in the HongHu dataset, and the second row shows the results for corn in the LongKou dataset.}
    \label{FCNs_results}
    \end{figure*}

Two key conclusions can be drawn from Table~\ref{FocalOneClassRiskEstimator}:
(1) The distribution-rebalancing strategy is an effective way to solve the problem of distribution imbalance.
The weight of the class risk ${\alpha}_{p}$ is analyzed when ${\gamma}$ is fixed at 0 in Table~\ref{FocalOneClassRiskEstimator}(a), and reliable one-class classification results can be obtained when $\alpha \geq 0.2$.
These results show that the ratio of the weight of the risk between the positive and negative classes may not have to be limited to $\pi_{p}/(1-\pi_{p})$, and a reliable HSI one-class classifier can also be obtained from the conformance-based negative risk.
We believe that the distribution rebalancing acts as a regularization in one-class classification because we force the model to assign approximate weights to the classes with different distribution frequencies.
We note that the average F1-score decreases when $\alpha=0.4$.
What is more, we note that this is due to the larger standard deviation of the results when $\alpha=0.4$.
When $\gamma$ is added, this shortcoming is compensated, and the standard deviation of the results decreases under certain combinations of $\alpha$ and $\gamma$.
For example, the results have a lower standard deviation when $\alpha=0.2$ and $\gamma=0.5$ or $\alpha=0.3$ and $\gamma=0.1$.
(2) The second conclusion is that the dynamic weighting mechanism for the positive samples can further improve the performance of the \emph{One-Class Risk Estimator}, thanks to the attention paid to the hard positive samples, and the dynamic weighting mechanism has the ability to reduce the standard deviation.
From Table~\ref{FocalOneClassRiskEstimator}(b) to Table~\ref{FocalOneClassRiskEstimator}(e), the F1-score increases when an appropriate value of $\gamma$ is added, and the standard deviation decreases.
As shown in Table ~\ref{FocalOneClassRiskEstimator}, the best average F1-score for the HongHu dataset is obtained at ${\alpha}_{p}$=0.3 and ${\gamma}$=0.1, and the best average F1-score is 95.61.
However, different combinations of $\alpha$ and $\gamma$ give similar results.
For example, the F1-scores with $\alpha=0.2$ and $\gamma=0.5$ are similar to the F1-scores with $\alpha=0.3$ and $\gamma=0.1$, which shows that the proposed positive representation enhancement strategy is robust and does not depend on a unique combination of $\alpha$ and $\gamma$.
From the risk curves~(Fig.~\ref{Risk_Change}), it is evident that the positive data were better fitted after the positive representation enhancement strategy was applied.

\subsection{Analysis of Different Positive Training Samples}

Extensive experiments were conducted to study the impact of the number of training samples in one-class classification methods.
The classes "cotton", "rape" and "carrot" in the HongHu dataset were selected as examples, and the results are shown in Fig.~\ref{training_samples_results}.
For fair comparison, the number of unlabeled samples are fixed at 4000.
Compared to other one-class classification methods, \textit{HOneCls} achieved the most robust results across different positive training samples and also performed well in few-shot learning scenario (5-shot positive training samples).

\subsection{Comparison between Class Imbalance and Distribution Imbalance}

Extensive experiments of carrot in HongHu dataset were conducted to demonstrate the difference between class imbalance and distribution imbalance.
The cause of class imbalance is the imbalance in the number of samples between positive and negative classes, therefore, adding more positive data can solve the problem of class imbalance at its root.
Although the number of positive training samples was increased from 100 to 3000 in the unbiased risk estimator (with the number of unlabeled samples fixed at 4000), the F1-score remained consistently 0 (Table~\ref{class_imb_dis_imb_f1}).
In other words, increasing the number of positive samples does not alleviate the problem of distribution imbalance, therefore, class imbalance and distribution imbalance are two different problems.

\subsection{Analysis of Different FCNs}
We used the mainstream FCNs (PSPNet, UNet, DeepLab-v3, DeepLab-v3+, and HRNet) as global spectral-spatial feature extractors with the proposed \emph{One-Class Risk Estimator} for the one-class classification task.
Due to the lack of boundary information in the sparse labels, the mainstream FCNs show the phenomenon of boundary distortion with \emph{HOneCls} (Fig.~\ref{FCNs_results}).
The average F1-scores of the different FCNs are listed in Table~\ref{FCNs_f1}, where the best results are obtained by the proposed \emph{FreeOCNet} extractor.
The FCN proposed in this paper, i.e., \emph{FreeOCNet}, avoids the boundary distortion caused by the sparse labels.
The feature maps with the same size in the encoder and decoder are fused to maintain better spatial details, and the approach of point-wise addition obtains better results than the approach of concatenation.
 
\begin{figure*}[!t]
    \centering
    \includegraphics[width=0.96\textwidth]{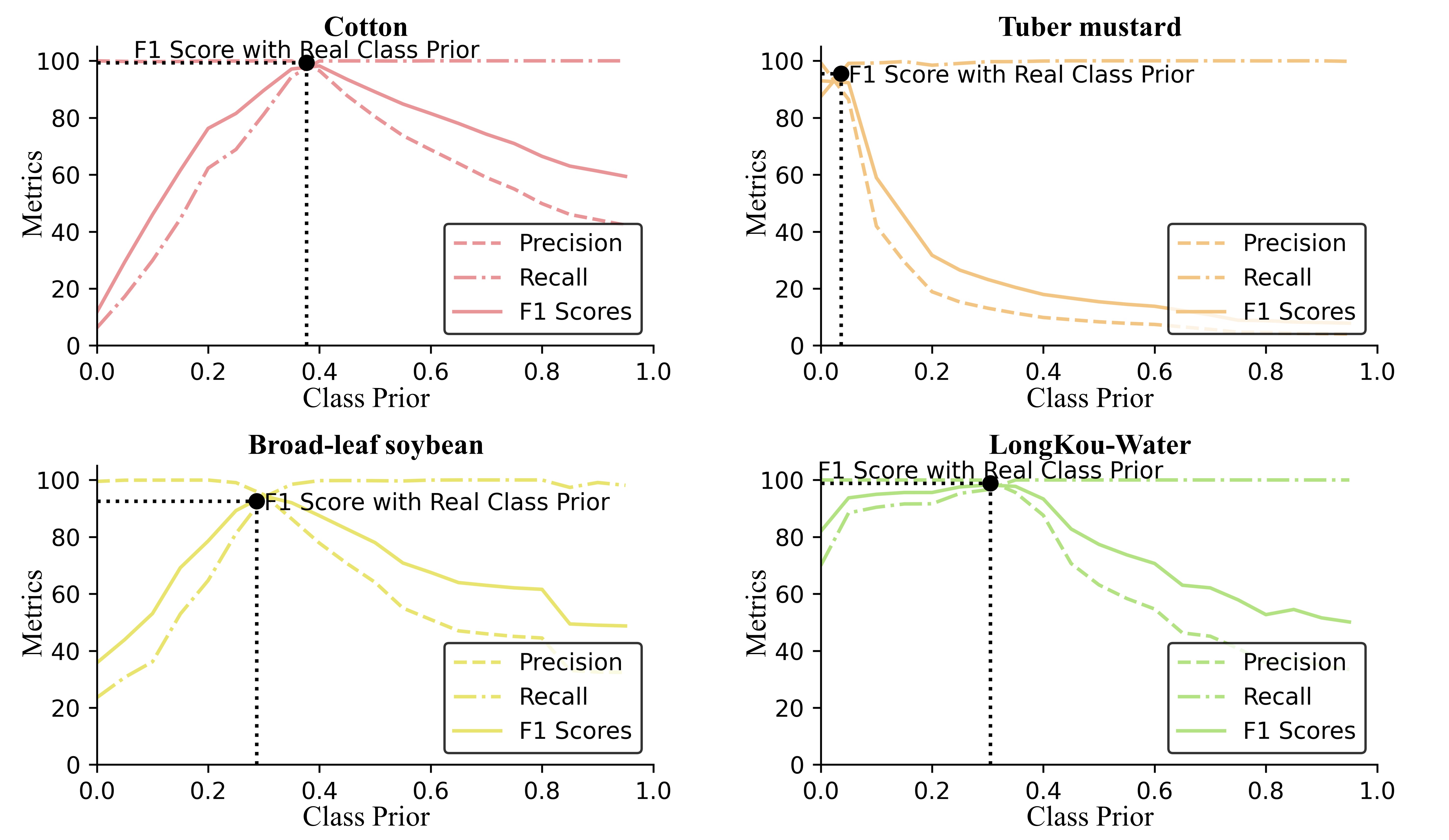}
    \caption{Analysis of the class prior. The dots indicate an exact class prior.}
    \label{class_prior_analysis}
    \end{figure*}

\begin{table}[!b]
        \caption{The average F1 scores of unbiased risk estimator and one-class risk estimator of carrot in HongHu. The mean and standard deviation are reported.}
        \label{class_imb_dis_imb_f1}
        \centering
        \resizebox{.95\columnwidth}{!}{
        \begin{tabular}{ccccc}
        \toprule
        Risk estimator   & \multicolumn{3}{c}{Unbiased risk estimator}                                        & One-class risk estimator \\
     \# Positive samples & 100        & 1000      & 3000       & 100                     \\ \midrule
        F1-scores        & 0(0)       & 0(0)      & 0(0)       & 93.58(0.82)             \\ \bottomrule
    \end{tabular}}
    \end{table}    

\begin{table}[!b]
    \tiny
    \caption{The average F1 scores of the different FCNs. The mean and standard deviation are reported.}
    \label{FCNs_f1}
    \centering
    \resizebox{.95\columnwidth}{!}{
    \begin{tabular}{cccc}
    \toprule
    Networks    & HongHu               & LongKou              & HanChuan             \\ \midrule
PSPNet      & 94.47(1.02)          & 97.09(0.33)          & 91.86(2.01)          \\
UNet        & 89.63(3.52)          & 90.43(3.66)          & 80.61(3.91)          \\
DeepLab V3  & 93.51(0.37)          & 94.41(0.50)          & 87.78(1.07)          \\
DeepLab V3+ & 94.20(0.94)          & 96.60(0.82)          & 89.82(1.69)          \\
HRNet       & 93.42(0.39)          & 95.81(0.47)          & 88.28(0.53)          \\
FreeOCNet   & \textbf{95.61(0.62)} & \textbf{98.07(0.68)} & \textbf{93.63(1.53)} \\ \bottomrule
    \end{tabular}}
    \end{table}

\subsection{Analysis of the Role of Class Prior in the Adaptive Flood Level}
In this part, the role of class prior for HSI one-class classification is analyzed.
The influence of the class prior on the detection results was analyzed at an interval of 0.05.
The results are shown in Fig.~\ref{class_prior_analysis}.

The dots in Fig.~\ref{class_prior_analysis} are the results of the approximate real class prior and its F1 score.
We found that the class prior controls the precision and recall of the model.
When the input class prior is smaller than the real class prior, the model obtains a high precision, and the recall increases with the increase of the class prior.
When the input class prior is larger than the real class prior, the model obtains a high recall, but with the increase of the class prior, the model precision gradually declines.
What is more, in the case of inputting a real class prior, the detection result will generally be better, but the input of a real class prior may not obtain the best result, as with the broad-leaf soybean class in the LongKou dataset.
It can be seen from Fig.~\ref{class_prior_analysis} that reliable results can still be obtained even when there is a 0.05 error in the estimated class prior.
In conclusion, the class prior in adaptive ``flood level'' (${\pi}_{p}\widehat{R}_{p}^{-}(f)$) controls the fitting degree of the model to the positive class.

\section{Conclusion}
In this paper, we have proposed a novel framework for the HSI one-class classification task---\emph{HOneCls}---which pushs the HSI classification methods from multi-class classification to one-class classification and reduces the annotation work in binary classification from the perspective of the class system.
In the \emph{HOneCls} framework, the \emph{One-Class Risk Estimator} has also been proposed in this paper, which suitable for the characteristics of distribution overlap and distribution imbalance of HSI.
The conformance-based negative risk estimator is introduced to \emph{One-Class Risk Estimator} to calculate a high quality risk for the negative class without negative data, and then the positive representation enhancement strategy is introduced to solve the under-fitting problem of the positive class.
The detection results were verified on three challenging aerial hyperspectral classification datasets, where the proposed \emph{HOneCls} framework showed a significant improvement over the other methods.

\bibliographystyle{./IEEEtran}
\bibliography{./IEEEabrv,./HOneCls_ref}

\begin{IEEEbiography}[{\includegraphics[width=1in,height=1.25in,clip,keepaspectratio]{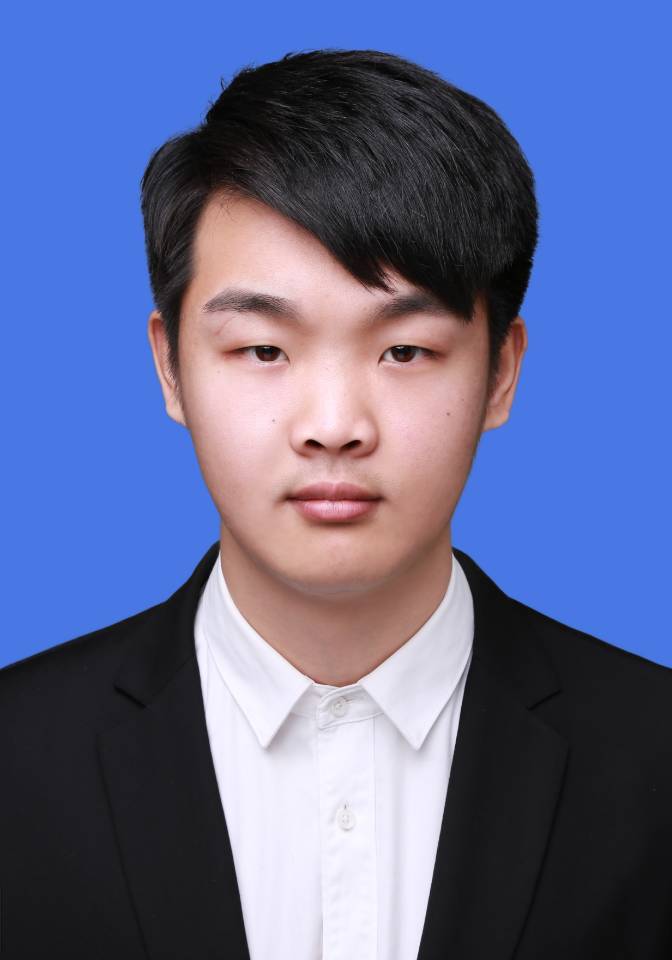}}]{Hengwei Zhao}
received the B.S. degree in surveying and mapping engineering from the School of Resources and Civil Engineering, Northeastern University, ShenYang, China, in 2019.
He is currently pursuing the Ph.D. degree in photogrammetry and remote sensing with the State Key Laboratory of Information Engineering in Surveying, Mapping and Remote Sensing, Wuhan University, Wuhan.

His major research interests are weakly supervised remote sensing image processing.
\end{IEEEbiography}
\vspace{-10mm}
\begin{IEEEbiography}[{\includegraphics[width=1in,height=1.25in,clip,keepaspectratio]{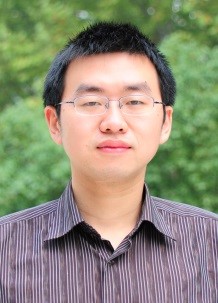}}]{Yanfei Zhong}(Senior Member, IEEE)
received the B.S. degree in information engineering and the Ph.D. degree in photogrammetry and remote sensing from Wuhan University, China, in 2002 and 2007, respectively.

Since 2010, He has been a Full Professor with the State Key Laboratory of Information Engineering in Surveying, Mapping and Remote Sensing (LIESMARS), Wuhan University, China.
He organized the Intelligent Data Extraction, Analysis and Applications of Remote Sensing (RSIDEA) research group.
He has published more than 100 research papers in international journals, such as Remote Sensing of Environment, ISPRS Journal of Photogrammetry and Remote Sensing, and IEEE TRANSACTIONS on GEOSCIENCE and REMOTE SENSING.
His research interests include hyperspectral remote sensing information processing, high-resolution remote sensing image understanding, and geoscience interpretation for multisource remote sensing data and applications.

Dr. Zhong is a Fellow of the Institution of Engineering and Technology (IET).
He was a recipient of the 2016 Best Paper Theoretical Innovation Award from the International Society for Optics and Photonics (SPIE).
He won the Second-Place Prize in the 2013 IEEE GRSS Data Fusion Contest and the Single-view Semantic 3-D Challenge of the 2019 IEEE GRSS Data Fusion Contest, respectively.
He is currently serving as an Associate Editor for IEEE JOURNAL of SELECTED TOPICS in APPLIED EARTH OBSERVATIONS and REMOTE SENSING and the International Journal of Remote Sensing.
\end{IEEEbiography}
\vspace{-10mm}
\begin{IEEEbiography}[{\includegraphics[width=1in,height=1.25in,clip,keepaspectratio]{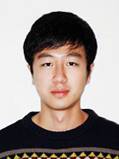}}]{Xinyu Wang}(Member, IEEE)
received the B.S. degree in photogrammetry and remote sensing, from the School of RemoteSensing and Information Engineering in 2014, and the Ph.D. degree from communication and information systemsin the State Key Laboratory of Information Engineering in Surveying, Mappingand Remote Sensing (LIESMARS) in 2019, from Wuhan University, Wuhan, China.
Since 2019, he has been an associate research fellow at the School of Remote Sensing and Information Engineering, Wuhan University, Wuhan.

His major research interests range include hyperspectral unmixing, detection, restoration and tracking.
\end{IEEEbiography}
\vspace{-10mm}
\begin{IEEEbiography}[{\includegraphics[width=1in,height=1.25in,clip,keepaspectratio]{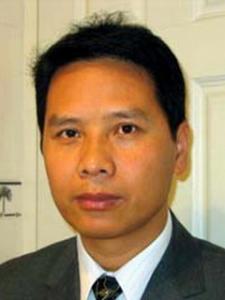}}]{Hong Shu}
received the bachelor’s from the Mathematics Department of Central China Normal University, the master’s degree from the Computer Department, Wuhan University, China, and the Ph.D. degree from the Department of Photogrammetry and Remote Sensing, State Key Laboratory of Information Engineering in Surveying, Mapping and Remote Sensing, Wuhan University.
He is currently a Professor of Geoinformatics with the State Key Laboratory for Information Engineering in Surveying, Mapping and Remote Sensing, Wuhan University.

His research focuses on spatio-temporal geographical computation, mainly including spatio-temporal geostatistics, computational geometry and topology, spatio-temporal scale analysis, land surface data assimilation, and space-time common sense reasoning.
He is a Principal Investigator of more than twenty research projects and an author of approximately 100 academic papers.
\end{IEEEbiography}

\clearpage
{\appendices
\section*{Appendix A. Proof of the bias and consistency}
The probability density functions of $\mathcal{X}_p$ and $\mathcal{X}_u$ can be denoted as $p_p(\mathcal{X}_p)=p_p(\boldsymbol{x}^p_1) \cdots p_p(\boldsymbol{x}^{p}_{n_{p}})$ and $p_u(\mathcal{X}_u)=p(\boldsymbol{x}^u_1) \cdots p(\boldsymbol{x}^{u}_{n_{u}})$, respectively.
The joint cumulative distribution function of $(\mathcal{X}_p, \mathcal{X}_u)$ can then be denoted as
\begin{equation}
\nonumber
\label{joint_cumulative_distribution_function}
F(\mathcal{X}_p, \mathcal{X}_u)=F_p(\mathcal{X}_p) \cdot F_u(\mathcal{X}_u),
\end{equation}
where $F_p(\mathcal{X}_p)$ and $F_u(\mathcal{X}_u)$ are, respectively, the cumulative distribution functions of the PU data.

The probability measure of $\mathcal{S}^-(f)$ can be defined as
\begin{equation}
\nonumber
\label{probability_measure}
P(\mathcal{S}^-(f))= \int_{(\mathcal{X}_p, \mathcal{X}_u) \in \mathcal{S}^-(f)} dF(\mathcal{X}_p, \mathcal{X}_u).
\end{equation}
For $P(\mathcal{S}^-(f))$, we have the following mathematical transformation based on the assumptions:
\begin{equation}
\nonumber
\begin{aligned}
P(\mathcal{S}^-(f)) &= P(\widehat{R}^-_u(f)-\pi_p\widehat{R}^-_p(g)<0)\\ &\leq P(\widehat{R}^-_u(f)-\pi_p\widehat{R}^-_p(g) \leq (1-\pi_p)R^-_n(f)-\alpha)\\ &=P((1-\pi_p)R^-_n(f)-(\widehat{R}^-_u(f)-\pi_p\widehat{R}^-_p(g)) \geq \alpha).
\end{aligned}
\end{equation}

To further prove Lemma 1, we apply McDiarmid's inequality to the probability expression.
Given the assumption that $0 \leq l(f(\boldsymbol{x}), \pm 1) \leq \mathcal{C}_l$, there will be no more than $\mathcal{C}_l/n_p$ in the change of $\widehat{R}^-_p(f)$ when a sample of $\mathcal{X}_p$ is replaced, and no more than $\mathcal{C}_l/n_u$ in the change of $\widehat{R}^-_u(f)$ when a sample of $\mathcal{X}_u$ is replaced.
According to McDiarmid's inequality, for $\alpha$ we have
\begin{equation}
\nonumber
\label{McDiarmid's_inequality}
\begin{aligned}
&P((1-\pi_p)R^-_n(f)-(\widehat{R}^-_u(f)-\pi_p \widehat{R}^-_p(f)) \geq \alpha) \\
&=P(E[\widehat{R}^-_u(f)-\pi_p \widehat{R}^-_p(f)]-(\widehat{R}^-_u(f)-\pi_p \widehat{R}^-_p(f))\geq \alpha)\\
&\leq exp(-\frac{2\alpha^2}{n_p(C_l\pi_p/n_p)^2+n_u(C_l/n_u)^2}) \\ &=exp(-\frac{2\alpha^2/{C_l}^2}{\pi^2_p/n_p+1/n_u}).
\end{aligned}
\end{equation}
Thus, Lemma 1 can be proved:
\begin{equation}
\begin{aligned}
P(\mathcal{S}^-(f)) \le exp(-2(\alpha/C_l)^2/(\pi_p^2/n_p+1/n_u)).
\end{aligned}
\end{equation}
Then, to prove Theorem 1, we have two steps.
STEP 1 prove (\ref{theorem1_bias}) with the definition and Lemma 1; STEP 2: prove the further conclusion (\ref{theorem1_consistency}) with (\ref{lemma1}), the absolute value inequality and McDiarmid's inequality.

\noindent STEP 1:

Based on $\widehat{R}^-_{oc-n}(f)-\widehat{R}^-_n(f)=0$ on $\mathcal{S}^+(f)$, we have
\begin{equation}
\label{e[r]-r}
\begin{aligned}
\mathbb{E}[\widehat{R}^-_{oc-n}(f)]-R^-_n(f) =\mathbb{E}[\widehat{R}^-_{oc-n}(f)-\widehat{R}^-_n(f)] \\ =\int_{(\mathcal{X}_p, \mathcal{X}_u) \in \mathcal{S}^-(f)} \widehat{R}^-_{oc-n}(f)-\widehat{R}^-_n(f)dF(\mathcal{X}_p, \mathcal{X}_u).
\end{aligned}
\end{equation}
Hence, the bias of $\widehat{R}^-_{oc-n}(f)$ is greater than zero if and only if $\mathcal{S}^-(f)$ is non-zero, since $\widehat{R}^-_{oc-n}(f) > \widehat{R}^-_n(f)\ on\ \mathcal{S}^-(f)$, which means we gain the left part of (\ref{theorem1_bias}).
Based on (\ref{e[r]-r}) and Lemma 1, the exponential decay of the bias can be obtained:
\begin{equation}
\label{eq:1}
\begin{aligned}
& \mathbb{E}[\widehat{R}^-_{oc-n}(f)]-R^-_n(f)\\
&=\int_{(\mathcal{X}_p, \mathcal{X}_u) \in \mathcal{S}^-(f)} \widehat{R}^-_{oc-n}(f)-\widehat{R}^-_n(f)dF(\mathcal{X}_p, \mathcal{X}_u) \\
& \leq sup_{(\mathcal{X}_p, \mathcal{X}_u) \in \mathcal{S}^-(f)}\{\widehat{R}^-_{oc-n}(f)-\widehat{R}^-_n(f)\}\cdot P(\mathcal{S}^-(f))
\\& \leq \frac{2}{1-\pi_p}\pi_p\mathcal{C}_l\Delta_f
\end{aligned}
\end{equation}

\noindent STEP 2:

To gain the further conclusion, consider the deviation bound of $\widehat{R}^-_{oc-n}(f)$:
\begin{equation}
\nonumber
\begin{aligned}
& |\widehat{R}^-_{oc-n}(f)-R^-_n(f)|\\
& \leq |\widehat{R}^-_{oc-n}(f)-\widehat{R}^-_n(f)| + |\widehat{R}^-_n(f)-R^-_n(f)|\\
& \leq |\widehat{R}^-_{oc-n}(f)-\widehat{R}^-_n(f)| + |\widehat{R}^-_n(f)-\mathbb{E}[\widehat{R}^-_n(f)]|.
\end{aligned}
\end{equation}
Still, we apply McDiarmid's inequality to the part $|\widehat{R}^-_{n}(f)-\mathbb{E}[\widehat{R}^-_{n}(f)]|$.
There will be no more than $\pi_pC_l/(n_p-\pi_pn_p)$ in the change of $\widehat{R}^-_{n}(f)$ when a sample of $\mathcal{X}_p$ is replaced, and no more than $C_l/(n_u-\pi_pn_u)$ in the change of $\widehat{R}^-_{n}(f)$ when a sample of $\mathcal{X}_u$ is replaced.

\noindent For any $\epsilon > 0$:
\begin{equation}
\nonumber
\begin{aligned}
& P(|\widehat{R}^-_n(f)-\mathbb{E}[\widehat{R}^-_n(f)]|\geq \epsilon)\\
& \leq 2exp(-\frac{2\epsilon^2}{n_p(\pi_pC_l/(n_p-\pi_pn_p))^2+n_u(C_l/(n_u-\pi_pn_u))^2})\\
& \leq 2exp(-\frac{2\epsilon^2(1-\pi_p)^2n_un_p}{{C_l}^2(\pi_p^2n_u+n_p)}).
\end{aligned}
\end{equation}

Let $\sigma$ equal the right side of the above formula, and after transformation we have
\begin{equation}
\nonumber
\begin{aligned}
& \epsilon = \frac{1}{1-\pi_p}\sqrt{\frac{ln(2/\sigma) C_l^2}{2}(\frac{\pi_p^2}{n_p}+\frac{1}{n_u}})\\
& \leq C_\sigma(\frac{\pi_p}{\sqrt {n_p}} + \frac{1}{\sqrt {n_u}}).
\end{aligned}
\end{equation}
Thus, the result is that, with the probability of at least $1-\sigma$:
\begin{equation}
\nonumber
\begin{aligned}
|\widehat{R}^-_{n}(f)-\mathbb{E}[\widehat{R}^-_{n}(f)]|&\leq \epsilon\\
& \leq C_{\sigma} (\frac{\pi_p}{\sqrt {n_p}} + \frac{1}{\sqrt {n_u}})
\\ & = C_{\sigma}\cdot\mathcal{X}_{n_p,n_u}.
\end{aligned}
\end{equation}
Given that $|\widehat{R}^-_{oc-n}(f)-\widehat{R}^-_n(f)|>0$ with the probability of at most $\Delta_f$, we can prove (\ref{theorem1_consistency}):

\noindent For any $\sigma>0$, with the probability of at least $1-\sigma-\Delta_f$
\begin{equation}
\nonumber
\begin{aligned}
|\widehat{R}^-_{oc-n}(f)-R^-_n(f)|&\leq 0 +  |\widehat{R}^-_n(f)-\mathbb{E}[\widehat{R}^-_n(f)]|\\
&\leq C_{\sigma}\cdot\mathcal{X}_{n_p,n_u}
\end{aligned}
\end{equation}
The proof of Theorem 1 is then finished.

\section*{Appendix B. The Results of Precision/Recall}

The precision and recall of HongHu, LongKou and HanChuan datasets are shown in Table~\ref{HongHu_p_r}, Table~\ref{LongKou_p_r}, and Table~\ref{HanChuan_p_r}, respectively.
The main bottleneck for biased classifiers, post-calibration threshold classifiers and unbiased risk estimation-based classifiers are the inability to obtain high precision and recall on most tasks at the same time.
Generally, high precision with low recall means that the classifier tends to overfit the positive data, resulting in only samples that are extremely similar to the positive training data being correctly classified; high recall with low precision means that the classifier tends to underfit the positive data, resulting in a large number of negative samples being misclassified as positive.
In particularly, if both precision and recall are 0, it means that the classifier completely underfits the positive data, resulting in no samples being classified as positive.
Only the proposed \emph{HOneCls} enables the classifier to find a balance between the overfitting and underfitting of positive data in all tasks, and obtains high precision and recall at the same time.

\begin{table*}[!t]
    \tiny
    \caption{The Precision/Recall for the HongHu dataset}
    \label{HongHu_p_r}
    \centering
    \resizebox{.98\textwidth}{!}{
        \begin{tabular}{cccccccccccc}
            \toprule
            Class                    & Class prior & OCSVM       & BSVM        & PUL         & PBL         & DeepPUL     & DeepPBL     & BCE         & DOCC        & absNegative & HOneCls     \\ \midrule
            cotton                   & 0.3769      & 99.30/43.33 & 98.03/79.01 & 90.87/92.15 & 91.37/89.90 & 99.83/9.13  & 99.84/9.09  & 99.94/5.28  & 99.29/99.09 & 99.21/99.41 & 99.26/99.41 \\
            rape                     & 0.1317      & 92.82/44.28 & 92.70/76.50 & 80.97/31.11 & 82.19/27.39 & 97.87/53.10 & 98.01/52.91 & 99.69/47.83 & 97.49/92.38 & 99.39/97.77 & 99.26/97.93 \\
            chinese cabbage          & 0.0544      & 55.61/56.17 & 56.89/75.65 & 71.70/31.08 & 73.87/25.07 & 75.61/87.39 & 76.78/86.63 & 94.23/79.30 & 35.15/33.30 & 51.98/55.47 & 86.65/94.91 \\
            tuber mustard            & 0.0367      & 11.37/37.07 & 25.13/65.99 & 73.91/14.57 & 71.17/6.05  & 83.30/57.15 & 84.19/55.77 & 88.14/49.74 & 0.00/0.00   & 0.00/0.00   & 98.83/92.43 \\
            cabbage                  & 0.0319      & 93.93/47.06 & 96.72/88.80 & 97.04/28.92 & 98.16/20.76 & 99.56/74.05 & 99.70/72.89 & 99.87/69.68 & 0.00/0.00   & 20.00/17.48 & 99.92/96.92 \\
            brassica parachinensis   & 0.0194      & 19.93/54.32 & 22.65/85.04 & 72.55/43.07 & 80.11/34.84 & 93.75/84.53 & 94.65/82.60 & 95.34/80.72 & 0.00/0.00   & 0.00/0.00   & 91.44/97.91 \\
            white radish             & 0.0119      & 15.28/47.08 & 41.14/73.54 & 88.44/26.62 & 94.51/24.06 & 95.35/81.70 & 97.02/78.33 & 96.93/78.61 & 0.00/0.00   & 0.00/0.00   & 98.06/90.84 \\
            carrot                   & 0.0102      & 25.26/45.94 & 30.59/82.23 & 46.27/53.69 & 52.07/43.00 & 73.43/92.37 & 79.38/90.63 & 77.94/91.16 & 0.00/0.00   & 0.00/0.00   & 90.68/96.72 \\ \midrule
            Average Precision/Recall &             & 51.69/46.91 & 57.98/78.34 & 77.72/40.15 & 80.43/33.88 & 89.84/67.43 & 91.20/66.11 & 92.76/62.79 & 28.99/28.10 & 33.82/33.77 & 95.51/95.88 \\ \bottomrule
            \end{tabular}}
    \end{table*}
    
    \begin{table*}[!t]
    \tiny
    \caption{The Precision/Recall for the LongKou dataset}
    \label{LongKou_p_r}
    \centering
    \resizebox{.98\textwidth}{!}{
        \begin{tabular}{cccccccccccc}
            \toprule
            Class                    & Class prior & OCSVM        & BSVM         & PUL          & PBL         & DeepPUL     & DeepPBL     & BCE         & DOCC        & absNegative  & HOneCls     \\ \midrule
            water                    & 0.3048      & 100.00/52.44 & 100.00/79.13 & 100.00/64.71 & 99.99/64.43 & 99.48/68.92 & 99.49/68.76 & 99.72/49.46 & 99.99/97.71 & 100.00/93.96 & 99.99/97.81 \\
            broad-leaf soybean       & 0.2873      & 94.76/47.84  & 96.58/65.59  & 82.07/79.48  & 83.51/76.37 & 99.57/11.25 & 99.58/11.19 & 99.84/7.84  & 94.86/86.84 & 95.62/96.14  & 95.74/89.68 \\
            corn                     & 0.1569      & 99.32/51.68  & 98.56/89.79  & 90.96/71.29  & 91.86/67.26 & 99.81/19.90 & 99.81/19.73 & 99.92/14.99 & 99.97/99.34 & 99.33/99.93  & 99.97/99.20 \\
            rice                     & 0.0539      & 99.81/48.46  & 99.52/90.80  & 96.95/65.90  & 97.61/52.06 & 99.89/44.65 & 99.90/43.80 & 99.92/38.10 & 79.97/68.88 & 79.99/79.31  & 99.86/99.99 \\
            sesame                   & 0.0138      & 16.32/47.87  & 55.32/91.47  & 95.61/50.02  & 97.35/43.20 & 97.40/94.58 & 98.26/92.59 & 98.35/92.30 & 0.00/0.00   & 0.00/0.00    & 99.97/98.81 \\ \midrule
            Average Precision/Recall &             & 82.04/49.66  & 90.00/83.36  & 93.12/66.08  & 94.07/60.66 & 99.23/47.86 & 99.41/47.21 & 99.55/40.54 & 74.96/70.55 & 74.99/73.87  & 99.10/97.10 \\ \bottomrule
            \end{tabular}}
    \end{table*}

    \begin{table*}[!t]
        \tiny
        \caption{The Precision/Recall for the HanChuan dataset}
        \label{HanChuan_p_r}
        \centering
        \resizebox{.98\textwidth}{!}{
            \begin{tabular}{cccccccccccc}
                \toprule
                Class                    & Class prior & OCSVM       & BSVM        & PUL         & PBL         & DeepPUL     & DeepPBL     & BCE         & DOCC        & absNegative & HOneCls     \\ \midrule
                water                    & 0.2045      & 97.13/44.26 & 99.40/89.83 & 94.38/40.59 & 94.57/40.95 & 85.92/96.36 & 86.13/96.32 & 97.20/77.62 & 99.97/95.94 & 99.95/96.00 & 99.97/95.77 \\
                strawberry               & 0.1213      & 85.99/56.05 & 76.80/84.41 & 88.38/3.85  & 88.67/2.15  & 89.57/76.07 & 89.94/75.72 & 97.19/63.60 & 99.00/78.99 & 99.11/82.87 & 99.62/88.42 \\
                cowpea                   & 0.0617      & 27.50/46.14 & 31.79/79.25 & 93.15/2.26  & 93.21/2.28  & 99.40/27.60 & 99.44/27.13 & 99.61/23.10 & 0.00/0.00   & 0.00/0.00   & 98.22/76.04 \\
                road                     & 0.0503      & 31.85/45.46 & 55.08/81.73 & 66.94/1.57  & 66.78/1.51  & 94.49/66.84 & 95.04/65.63 & 97.16/57.04 & 0.00/0.00   & 0.00/0.00   & 96.85/86.83 \\
                soybean                  & 0.0279      & 49.41/50.31 & 30.80/92.97 & 71.93/23.03 & 74.88/15.52 & 82.56/70.96 & 83.01/69.66 & 84.05/66.33 & 0.00/0.00   & 0.00/0.00   & 96.27/97.80 \\
                watermelon               & 0.0123      & 8.09/51.66  & 10.14/72.68 & 42.67/3.74  & 44.10/1.36  & 94.31/75.65 & 95.25/73.39 & 95.23/73.46 & 0.00/0.00   & 0.00/0.00   & 91.49/90.90 \\
                water spinach            & 0.0033      & 13.80/50.27 & 10.95/88.00 & 35.49/31.29 & 55.07/6.73  & 96.83/96.62 & 98.05/94.60 & 97.05/96.31 & 0.00/0.00   & 0.00/0.00   & 97.36/99.96 \\ \midrule
                Average Precision/Recall &             & 44.82/49.16 & 44.99/84.13 & 70.42/15.19 & 73.90/10.07 & 91.87/72.87 & 92.41/71.78 & 95.36/65.35 & 28.42/24.99 & 28.44/25.55 & 97.11/90.82 \\ \bottomrule
                \end{tabular}}
        \end{table*}
}
\end{document}